\documentclass[journal]{IEEEtran}

\usepackage{amsmath,amsfonts}
\usepackage{graphicx}
\graphicspath{{figures/}}
\usepackage{cite}
\usepackage{hyperref}
\usepackage{booktabs}
\usepackage{multirow}

\begin{document}

\title{Beyond Accuracy: Assessing Calibration of Geospatial
Foundation Models and Their Sensitivity to Distribution Shifts}

\author{Nils Lehmann,
        Jakob Gawlikowski,
        Burak Ekim,
        Isaac Corley,
        and Xiao Xiang Zhu%
        ~\IEEEmembership{Fellow, IEEE}%
        \thanks{Nils Lehmann and Xiao Xiang Zhu are with the Chair of Data
        Science in Earth Observation, Technical University of Munich (TUM),
        Germany (e-mail: n.lehmann@tum.de; xiaoxiang.zhu@tum.de).}%
        \thanks{Jakob Gawlikowski is with the Remote Sensing Technology
        Institute, German Aerospace Center (DLR), Germany (e-mail: jakob.gawlikowski@dlr.de).}%
        \thanks{Burak Ekim is with the Computer Science and Artificial Intelligence Laboratory (CSAIL), Massachusetts Institute of Technology (MIT), USA (email: burake@mit.edu).}%
        \thanks{Isaac Corley is with Taylor Geospatial, USA
        (e-mail: isaac.corley@taylorgeospatial.org).}%
}

\maketitle

\begin{abstract}
Geospatial Foundation Models (GeoFMs) are most commonly ranked and selected by accuracy on
standard benchmark conditions via averaged ranks. We show that this protocol is too narrow: the
promised deployment in critical EO tasks requires further angles of analysis, mainly
\emph{calibration}, the agreement between a model's confidence and its correctness. Across 16
frozen encoders, four classification and five segmentation datasets, and two orthogonal stress
axes (physically motivated distribution shift and training-data budget), every encoder degrades
as corruption intensifies, and the ranking moves with it: rank transfer decays with severity in
all three corruption families, and by severity grade 3 a clean-data ranking recovers little of
the shifted ranking. Across the four classification benchmarks, EO-pretrained and
ImageNet-pretrained encoders are indistinguishable on clean accuracy and clean calibration, and
EO pretraining provides no more stability under shift than ImageNet pretraining. Under shift the
GeoFMs drift further into overconfidence than the ImageNet-pretrained encoders, at every grade and
in every corruption family. A centered kernel alignment (CKA) analysis ties this to representational
rigidity: EO-pretrained embeddings move less under corruption while losing just as much task
information and remaining overconfident. We apply three commonly explored uncertainty quantification
methods and find that temperature scaling and deep ensembles cannot counteract the degradation, while a
Gaussian-process probe roughly halves ECE under severe cloud only by tripling it on clean data. In selective prediction experiments, we find
that confidence-based abstention cannot defer around confidently wrong predictions, and advocate
that benchmark rankings and evaluations should therefore operate across a multitude of conditions
and metrics to more holistically evaluate model development progress and close the gap to real
world deployment scenarios.
\end{abstract}


\begin{figure*}[t]
  \centering
  \includegraphics[width=\textwidth]{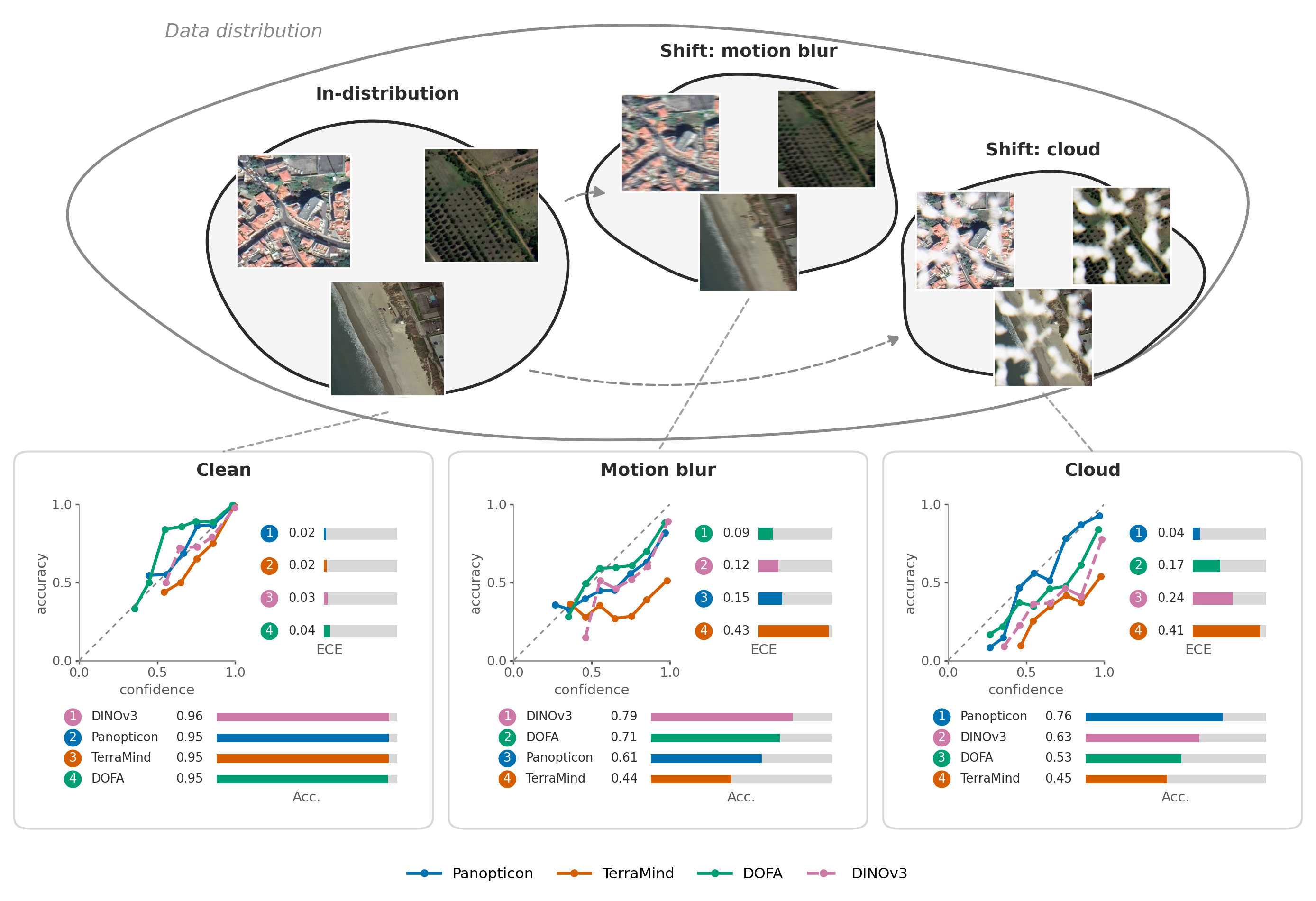}
  \caption{Conceptual overview. Frozen encoders are evaluated under physically simulated,
  test-time distribution shifts, shown as example shifts branching from in-distribution EO
  scenes (top). For one representative dataset (bottom), each shift is read out as a
  reliability curve (the dashed diagonal marks perfect calibration) with ECE and accuracy
  rankings. We show an illustrative subset of four encoders, three EO-pretrained and an
  ImageNet reference (DINOv3, dashed), not the full set of encoders evaluated in
  Section~\ref{sec:results}.
  Encoders that are nearly indistinguishable on clean data degrade in both accuracy and
  calibration under shift, and their rankings reshuffle.}
  \label{fig:hero}
\end{figure*}

\section{Introduction}
\IEEEPARstart{T}{he} term foundation model was introduced by Bommasani et
al.~\cite{bommasani2021opportunities} to describe a model that is trained on a vast corpus of
data to subsequently be adapted to many diverse downstream tasks. Following their success in
language modeling~\cite{brown2020language} as well as image~\cite{rombach2022high} and
video~\cite{ho2022video} generation, several dozen geospatial foundation models (GeoFMs) have
been proposed~\cite{huang2025survey}, motivated by the scale of available geospatial data and the
potential for improved label efficiency and downstream performance at lower fine-tuning
cost~\cite{zhu2026foundations}. Their defining promise is breadth: a single pretrained encoder
meant to serve a wide range of sensors, regions, imaging conditions, to enable or improve performance on various downstream tasks~\cite{zhu2026foundations}.

However, hat promise is difficult to check with current evaluation practices. Corley et al.~\cite{corley2026no}
recently demonstrated that, due to insufficient community standards and evaluation protocols,
practitioners are not yet able to holistically and fairly compare GeoFM performance on
downstream tasks. The disparity is visible across the literature: the same encoder on the same
benchmark is reported with different numbers in different papers, sometimes under the nominaly same evaluation protocol. Accuracy on clean benchmark conditions is also blind to what deployment conditions a model encounters, since safety-critical decision making~\cite{mehrtash2020confidence} and natural
distribution shifts of geography, sensors, and atmospheric conditions all change the input
distribution at test time~\cite{ovadia2019can, hendrycks2021many, zhu2026foundations}.

In order to more holistically evaluate GeoFM performance under these conditions, this study concerns test-time covariate shift. Distribution shift is commonly categorized
as semantic, covariate, or concept drift~\cite{moreno2012unifying, quinonero2009dataset,
tamang2025handling}; the shifts we consider perturb only the input while leaving the
labeling task intact, and are therefore covariate to the extent that the label remains
recoverable, in line with the common-corruption robustness
benchmarks~\cite{hendrycks2021many, ovadia2019can}. Shift also lies on a spectrum rather
than a dichotomy~\cite{doerksen2026earthshift}: applying an model to a distincly different
sensor or region is a far shift, whereas the same scenes and sensors observed through
atmospheric or sensor degradation is a near shift, and it is this near regime that we
apply. Figure~\ref{fig:hero} illustrates the resulting phenomenon: encoders
that are nearly indistinguishable on clean data degrade in both accuracy and calibration
under such shifts, and their rankings change.

In our evaluation we put particular emphasis on \emph{calibration} which describes the agreement between a model's
confidence and its correctness, and therefore whether a prediction can be acted on when its
accuracy cannot be checked directly \cite{guo2017calibration}. Calibration metrics are well established in the general machine learning (ML)
literature but have barely been applied in the geospatial domain, as
Section~\ref{sec:related_work} shows. In this regard, our contribution is a systematic assessment
of how sensitive GeoFM calibration is to distribution shift, across a diverse set of
classification and segmentation datasets (Table~\ref{tab:datasets}), following established
task-adaptation protocols.

Our central finding is one about \emph{evaluation}, not about any single model.
When we re-evaluate the same encoders under physically motivated EO distribution shifts and
across training-data budgets, model rankings shift substantially under most of
the conditions we test. The single number that is commonly reported, accuracy
on clean, full-data test sets, is not informative enough for indicating
trustworthy behavior. Rank transfer degrades systematically with corruption
severity in all three corruption families, and by severity grade 3 a clean-data
ranking recovers little of the shifted ranking. We make this concrete along two
orthogonal axes, distribution shift and training-data budget, use a centered kernel alignment (CKA) representational
analysis to assess feature sensitivity, and furthermore show that
common calibration methods do not fully resolve these issues. Concretely, we
contribute a corpus audit establishing that calibration is absent at every level
of the GeoFM evaluation ecosystem (Section~\ref{sec:related_work}); a systematic
assessment of the calibration of 16 frozen encoders under EO-specific corruptions
and across five training-data
budgets (Sections~\ref{sec:shift}--\ref{sec:budget}); a
representation-drift analysis that ties calibration failure to a
domain-conditioned pattern (Section~\ref{sec:mechanism}); and an
evaluation of post-hoc calibration, ensembles, and a Gaussian-process probe, none
of which fully restores calibration under shift. Furthermore, a
selective-prediction evaluation shows that confidence-based abstention cannot
defer around confidently wrong predictions: at the highest cloud severity, coverage is near
total and mean accuracy falls to 0.15, yet models remain overconfident
(Sections~\ref{sec:mitigations}--\ref{sec:selective}).

\section{Related Work}
\label{sec:related_work}

Reliable machine learning requires models that are not only accurate but
\emph{trustworthy}~\cite{mucsanyi2023trustworthy, liu2021trustworthy},
a property we define to encompass three distinct dimensions:
\emph{calibration} (whether predicted confidence matches empirical accuracy
\emph{in aggregate}~\cite{guo2017calibration}),
\emph{predictive uncertainty} (whether a model can identify \emph{which
individual inputs} it is ignorant about~\cite{gawlikowski2023survey}),
and \emph{robustness} (whether calibration and uncertainty \emph{continue to
hold under distribution shift}, so that a model's confidence degrades
gracefully alongside its accuracy~\cite{hendrycks2021many}).
These dimensions are related but not interchangeable: calibration is a
\emph{marginal} property that constrains confidence only on average and can be
satisfied by an uninformative predictor that outputs the base rate for every
input, whereas predictive uncertainty is a \emph{conditional} property that
asks whether the model knows \emph{where} it will be wrong. Neither is
guaranteed to survive distribution shift, since a calibration map fit on clean
in-distribution data need not transfer once the test distribution moves.
We measure the conditional, predictive-uncertainty dimension with per-sample
proper scoring rules (NLL and Brier score; Section~\ref{sec:metrics}), which,
unlike the binned and marginal ECE, are sensitive to the probability assigned
to each individual label, and with selective-prediction diagnostics
(Section~\ref{sec:selective_protocol}) that test whether confidence identifies
the inputs a model gets wrong.
Each is a mature subfield in the broader ML literature:
calibration evaluation is standard in medical image
analysis~\cite{dominique2025role,ju2024monica} and language
modeling~\cite{kadavath2022language,center2026benchmark},
and large-scale robustness benchmarks in the general ML domain routinely pair accuracy with calibration metrics~\cite{ovadia2019can, minderer2021revisiting,tao2024benchmark}. As we show below, none of these practices have transferred to GeoFM evaluation.

\paragraph{Evaluation practices for geospatial foundation models}
GeoFMs are commonly evaluated on dozens of datasets encompassing diverse tasks such as classification, segmentation, and
regression~\cite{zhu2026foundations}.
Community dataset suites such as GEO-Bench-1 and 2 ~\cite{lacoste2023geo, simumba2025geo} and
PANGAEA~\cite{Marsocci2024PANGAEAAG} propose standardization of these procedures, but recent
audits reveal that evaluation throughout the community is neither reproducible nor standardized: across
152 self proclaimed GeoFM papers, identical model--benchmark pairs disagree by more than ten
accuracy points in 46 cases, most papers use a unique pre-training
configuration, and a substantial fraction release no model
weights~\cite{corley2026no}.
The gap we identify, however, extends further than reproducibility.
We systematically audited this same corpus, as well as available benchmark frameworks, for evidence of
calibration, predictive uncertainty, and robustness evaluation.
The finding is consistent across all three strands:
\emph{(i)}~not one of the 152 GeoFM model papers reports a calibration metric
in the machine-learning sense (no ECE, reliability diagrams, or Brier score); every ``calibration'' mention refers
to radiometric sensor calibration;
\emph{(ii)}~neither GEO-Bench nor PANGAEA include calibration or predictive
uncertainty in their evaluation protocols;
\emph{(iii)}~dedicated EO robustness benchmarks (REOBench~\cite{li2026reobench}
and EarthShift~\cite{doerksen2026earthshift}, which evaluate GeoFMs across dozens of shift conditions) report exclusively accuracy metrics. These audits let us conclude that calibration is currently absent at every level of the GeoFM evaluation ecosystem.

\paragraph{Robustness and distribution shift}
A minority of GeoFM papers probe robustness, but nearly always reframe a
model's headline capability as a generalization test with respect to accuracy performance rather than assessing model confidence.
The most common form of generalization testing is cross-region and cross-sensor generalization: holding
out continents to measure geographic shifts~\cite{Klemmer2023SatCLIPGG},
transferring to sensors unseen during
pre-training~\cite{Prexl2024SenPaMAESP, Braham2024SpectralEarthTH},
evaluating any-sensor encoders on out-of-distribution bands and
resolutions~\cite{waldmann2025panopticon}, and constructing
generalization benchmarks with respect to region and sensors ~\cite{Gong2024CrossEarthGV, Marsocci2024PANGAEAAG}.
A second strand studies scale and resolution
shift~\cite{Reed2022ScaleMAEAS, Tang2024CrossScaleMA} and degradation under
cloud cover or sparse temporal sampling~\cite{Feng2025TESSERATE}.
Among GeoFMs, only HyperSIGMA reports a more classic robustness check:
adversarial attacks (FGSM, PGD) and input corruptions (JPEG, Gaussian
noise)~\cite{Wang2024HyperSIGMAHI}.
A more deployment-oriented line of work asks whether shift can be
\emph{detected} rather than merely tolerated: Ekim et al.~\cite{ekim2025distribution} propose TARDIS, a post-hoc OOD detector
based on internal activations, and evaluate it across 17 EO covariate and
semantic shift setups on EuroSAT and xBD. This extends the EO robustness
literature from accuracy under shift to explicit shift-awareness at inference
time, but still does not study whether model confidence remains calibrated
under those shifts.
Critically, none of these works, nor the dedicated benchmarks
REOBench~\cite{li2026reobench} and EarthShift~\cite{doerksen2026earthshift},
connect robustness to whether a model's \emph{confidence} degrades gracefully
alongside its accuracy.

\paragraph{Uncertainty and calibration}
Calibration and uncertainty quantification are recognized dimensions of
trustworthy ML~\cite{mucsanyi2023trustworthy, liu2021trustworthy}, yet they
are nearly absent from the GeoFM literature.
Only a small number of EO papers estimate predictive uncertainty directly.
Chen and Bruzzone~\cite{Chen2021SelfsupervisedRS} use a negative-log-likelihood training
objective~\cite{kendall2017uncertainties} to estimate aleatoric uncertainty in
a self-supervised change detector. More recently,
SHRUG-FM~\cite{gonzalezcalabuig2026shrugfm} combines geophysical input-space
OOD cues, embedding-space OOD cues, and task-level uncertainty in a
selective-prediction framework for burn-scar, flood, and landslide mapping.
This is closely related to deployment-time shift awareness, but its focus is
abstention and risk reduction on retained samples rather than calibration in
the sense of reliability diagrams, ECE, NLL, or Brier score.
AlphaEarth~\cite{alphaearth2025} includes an ``Uncertainty estimation''
section but reports bootstrap confidence intervals on evaluation metrics rather
than per-prediction model uncertainty.
Calibration in the machine-learning sense (dedicated metrics, reliability diagrams,
and relationship to accuracy) is reported by \emph{no} model in the
corpus we audit; the closest is an unquantified remark that one model's higher
softmax entropy yields ``more calibrated predictions''~\cite{Danish2025TerraFMAS}.
Frequent mentions of ``calibration'' elsewhere refer to radiometric sensor
calibration, and ``temperature scaling'' almost always denotes the
contrastive-loss temperature, not confidence calibration.
Outside geospatial work, post-hoc calibration methods such as temperature
scaling~\cite{guo2017calibration}, deep
ensembles~\cite{lakshminarayanan2017simple}, and conformal
prediction~\cite{Angelopoulos2021AGI} are well established, and calibration
under distribution shift is actively studied: Ovadia et al.~\cite{ovadia2019can} show that
temperature scaling, while effective in-distribution, frequently fails under
even moderate shift; Hendrycks et al.~\cite{hendrycks2021many} find that no single method
consistently improves robustness across shift types; and
Minderer et al.~\cite{minderer2021revisiting} and Tao et al.~\cite{tao2024benchmark} report that the
accuracy--calibration correlation holds mainly among high-performing models and
does not generalize across datasets.

\paragraph{Our work}
To the best of our knowledge, we provide the first systematic
assessment of GeoFM calibration and uncertainty under distribution shift.
The neglect of calibration is consistent across all three strands of prior
work: individual model papers, community evaluation frameworks, and dedicated
robustness benchmarks all focus exclusively on accuracy.
Unlike these works, we jointly evaluate calibration (ECE, NLL, Brier score,
reliability diagrams), uncertainty quantification methods (post-hoc temperature
scaling, deep ensembles, and a GP probe), and representation drift under EO-specific distribution shifts
across a diverse set of GeoFMs and downstream classification tasks
(Table~\ref{tab:datasets}) using a common protocol. Under this approach we aim to establish
a more holistic view of GeoFM performance that includes how well these models are calibrated and
how this perspective can further support real-world application cases, where predictive uncertainty
and model prediction confidence matter.

\section{Methodology}
\label{sec:methodology}

\subsection{Problem Setting}
\label{sec:problem_setting}

We study the predictive calibration of frozen geospatial encoders under linear
probing, for both classification and segmentation.
Given a pre-trained backbone $f_\theta: \mathcal{X} \to \mathbb{R}^D$ with fixed
weights $\theta$, we train a lightweight probing head on the extracted
embeddings: for classification, a linear probe
$g_\phi: \mathbb{R}^D \to \mathbb{R}^K$ producing per-image class logits, where
$K$ is the number of classes; for segmentation, a lightweight FPN-style
decoder head~\cite{lin2017feature} $g_\phi$ takes the backbone's feature maps
at $L=4$ coarse-to-fine depths and produces per-pixel class logits
$g_\phi(\{z_l\}_{l=1}^{L}) \in \mathbb{R}^{K \times H \times W}$ over the
$H \times W$ spatial grid of the input image.
For the classification datasets, we also fit UQ methods but the
backbone weights are never modified, mirroring the dominant GeoFM deployment
pattern of a shared backbone with lightweight task-specific heads.

The probe is held constant across all 16 encoders: the same head, the same
protocol, and a regularization strength selected on accuracy and never re-tuned
on calibration loss (Section~\ref{sec:probe}). A single absolute
calibration number therefore depends on the head, but the between-encoder
differences that our claims rest on hold the head fixed and so isolate the
encoder.

A model is \emph{calibrated} if $\Pr(Y = y \mid \hat{p}(x) = p) = p$ for all
$p \in [0, 1]$~\cite{guo2017calibration}.
We measure calibration through three metrics (Section~\ref{sec:metrics}) under
clean conditions and three geospatially motivated distribution shifts as well as varying training data budgets.

\subsection{Datasets}
\label{sec:datasets}

We evaluate on two common prediction tasks (Table~\ref{tab:datasets}).
For classification we use four single-label remote-sensing datasets (ADVANCE,
m-EuroSAT, RESISC45, and So2Sat) spanning diverse sensors, spatial
resolutions, and label spaces. 
For segmentation we use five datasets (CloudSEN12~\cite{aybar2022cloudsen12},
DynamicEarthNet~\cite{toker2022dynamicearthnet}, FLAIR\#2~\cite{garioud2023flair},
Fields of the World~\cite{kerner2025fields}, and SpaceNet2~\cite{vanetten2018spacenet}) covering cloud/shadow
masking, land cover, agricultural parcels, and building footprints across
Sentinel-2, aerial, and high-resolution commercial sensors.

\begin{table}[t]
\centering
\caption{Evaluation datasets. \emph{Top:} single-label classification canon
($N_\text{train}$ includes validation samples not reserved for calibration;
$N_\text{cal}=400$ for all datasets). \emph{Bottom:} dense-prediction
(segmentation) datasets. Segmentation does not use a held-out calibration
split or the train/validation merge described in
Section~\ref{sec:probe}---post-hoc calibration methods
(Section~\ref{sec:mitigations}) apply to classification only---so
$N_\text{val}$ is reported instead of $N_\text{cal}$. SpaceNet7 ($\dagger$)
is in the corruption sweep but excluded from the five-dataset segmentation
canon used throughout the analysis (see Appendix, \emph{Exclusion of
SpaceNet7}).}
\label{tab:datasets}
\begin{tabular}{lrrrl}
\toprule
\multicolumn{5}{l}{\textbf{Classification}} \\
Dataset & $N_\text{train}$ & $N_\text{cal}$ & $N_\text{test}$ & Classes \\
\midrule
ADVANCE~\cite{hu2020cross}      &  3{,}660 & 400 &  1{,}015 & 10 \\
m-EuroSAT~\cite{lacoste2023geo} &  2{,}600 & 400 &  1{,}000 & 10 \\
RESISC45~\cite{cheng2017remote} & 24{,}800 & 400 &  6{,}300 & 45 \\
So2Sat~\cite{zhu2020so2sat}     & 20{,}578 & 400 &    986   & 17 \\
\bottomrule
\end{tabular}

\vspace{0.6em}

\begin{tabular}{lrrrl}
\toprule
\multicolumn{5}{l}{\textbf{Segmentation}} \\
Dataset & $N_\text{train}$ & $N_\text{val}$ & $N_\text{test}$ & Task \\
\midrule
CloudSEN12~\cite{aybar2022cloudsen12}       & 4{,}000 & 535   & 975   & Cloud/shadow \\
DynamicEarthNet~\cite{toker2022dynamicearthnet} & 700 & 100   & 200   & Land cover \\
FLAIR\#2~\cite{garioud2023flair}            & 4{,}049 & 1{,}022 & 3{,}022 & Land cover \\
Fields of the World~\cite{kerner2025fields}   & 4{,}000 & 1{,}000 & 2{,}000 & Field parcels \\
SpaceNet2~\cite{vanetten2018spacenet}       & 5{,}186 & 1{,}461 & 2{,}961 & Buildings \\
SpaceNet7$^\dagger$~\cite{vanetten2018spacenet} & 3{,}500 & 652   & 1{,}152 & Buildings \\
\bottomrule
\end{tabular}
\end{table}

\subsection{Encoders}
\label{sec:encoders}

We evaluate 16 frozen encoders (Table~\ref{tab:roster}). Following prior
practice we can group them by \emph{pre-training domain}: GeoFMs
pre-trained on satellite or aerial imagery, general-purpose encoders
pre-trained on ImageNet, and a non-learned Empirical Random Convolutional
Feature (RCF) baseline~\cite{rolf2021generalizable, corley2024revisiting}.
We deliberately include general purpose image encoders to study whether EO-specific pre-trained models
are calibrated better than non-domain specific encoders.

As a finer descriptive tag we also record each encoder's pre-training
\emph{objective}, split by the type of target it predicts: EO reconstruction
(EO-recon), which predicts masked input content against a \emph{fixed} target
(raw pixels, a frozen tokenizer, or a fixed random projection); EO
self-distillation (EO-distill), which predicts against a \emph{co-evolving}
teacher; and, on the natural-image side, self-distillation (Nat-DINO),
supervised (Nat-supervised), or random (RCF). Several EO encoders are
hybrids: Clay adds a small DINOv2 teacher term to a pixel-reconstruction loss,
DOFA distills from ImageNet-pretrained features, and OlmoEarth adds an
instance-contrastive term, so this split is a soft descriptive tag rather than a
clean dichotomy, and we use it for exploratory grouping only, not as a headline
axis.

\begin{table}[t]
\centering
\caption{The 16 frozen encoders used in this study, grouped by pretraining
provenance (satellite/aerial vs.\ natural-image) and pretraining objective
(target type).}
\label{tab:roster}
\begin{tabular}{llll}
\toprule
Paradigm & Provenance & Encoders \\
\midrule
EO-recon       & GeoFM    & Clay~V1.5~\cite{clay2024}, Prithvi~V2~\cite{szwarcman2025prithvi}, DOFA~\cite{xiong2024neural}, \\
               &          & OLMo-Earth base/tiny~\cite{tseng2026olmoearth}, TerraMind~\cite{jakubik2025terramind} \\
EO-distill     & GeoFM    & Panopticon~\cite{waldmann2025panopticon}, \\
               &          & ViT-L DINOv3-Sat~\cite{simeoni2025dinov3} \\
Nat-DINO       & General  & ConvNeXt-L DINOv3~\cite{simeoni2025dinov3} \\
Nat-supervised & General  & ResNet-18/50~\cite{he2016resnet}, Swin-T~\cite{liu2021swin}, \\
               &          & ViT-B/ViT-L~\cite{dosovitskiy2020vit}, \\
               &          & MobileNetV3~\cite{howard2019mobilenet} \\
Random         & Baseline & RCF~\cite{rolf2021generalizable} \\
\bottomrule
\end{tabular}
\end{table}

All images are resized to $224 \times 224$ pixels with bilinear interpolation
and normalized channel-wise using per-band statistics computed from the
training set or alternatively the own model prescribed normalization scheme. ADVANCE and RESISC45 are RGB imagery only, so every encoder is
evaluated in RGB on these two datasets. For m-EuroSAT and So2Sat, which
provide multispectral bands, each encoder was probed in both RGB
and its native multispectral configuration during the accuracy-focused linear
probe sweep (Section~\ref{sec:probe}), and the higher-accuracy configuration
was carried forward for calibration evaluation.
Evaluating each encoder at its best band configuration is a deliberate choice, since this reflects how benchmarks commonly attempt to establish a best model. The calibration effect we report is nonetheless not an artifact of the band choice: on ADVANCE
and RESISC45, where every encoder receives identical RGB input and no band
configuration is selected, the EO-versus-general calibration gap under shift is
if anything larger (cloud severity~3 ECE $0.44$ vs.\ $0.25$), while clean ECE is
indistinguishable ($0.023$ vs.\ $0.024$; Table~\ref{app:encoder_severity_full},
RGB-only group rows).

\subsection{Linear Probe Training and Calibration Split}
\label{sec:probe}

For each encoder--dataset pair we extract embeddings from the frozen backbone
and train a multinomial linear probe:
\begin{equation}
    g_\phi(z) = \operatorname{softmax}(Wz + b),
    \quad z = f_\theta(x),
\end{equation}
where $W \in \mathbb{R}^{K \times D}$ and $b \in \mathbb{R}^K$.
The probe is optimized with L-BFGS and $\ell_2$ regularization; the
regularization strength $C = 1/\lambda$ is taken from an accuracy-focused
linear probe sweep, without re-tuning on
calibration loss, so that any observed miscalibration reflects the encoder
and probe rather than the regularization choice.

\paragraph{Calibration split.}
A stratified random split of 400 samples is held out from the validation set
and reserved exclusively for fitting temperature scaling parameters.
Critically, $\mathcal{D}_\text{cal}$ is drawn entirely from clean, validation data: $T^*$ is fit once and never re-fit or
otherwise exposed to corrupted inputs, mirroring the realistic deployment
setting in which corrupted data is not available at calibration time.
Once hyperparameters have been selected on validation-set performance, the remaining validation samples are merged into $\mathcal{D}_\text{train}$ for a final round of retraining. All reported metrics are computed on the held-out test set.

\paragraph{Segmentation probe.}
For dense prediction we keep the same frozen-backbone approach but replace
the linear probe with a lightweight FPN decoder~\cite{lin2017feature}: we
hook $L=4$ coarse-to-fine feature layers from the frozen backbone, project
each to a shared hidden dimension with a $1{\times}1$ lateral convolution,
merge them top-down via bilinear upsample-and-add, refine each level with a
$3{\times}3$ convolution, then concatenate all levels and classify with a
final $1{\times}1$ convolution before upsampling to the input resolution. As
in the classification setting, only the decoder is trained and the backbone
is never updated. Calibration is then measured per pixel (pixel-ECE)
alongside mean intersection-over-union (mIoU), so that the
accuracy--calibration contrast is directly comparable to the classification
setting.

\subsection{Uncertainty Quantification Methods}
\label{sec:uq_methods}

We evaluate three UQ strategies applied to the same frozen embeddings
and trained probe, of which only temperature scaling is post-hoc.
These UQ methods are fit and evaluated on the classification datasets only;
segmentation is assessed under distribution shift and training-budget
variation but not under the UQ methods.

\paragraph{Uncalibrated (baseline).}
The baseline applies softmax directly to the linear probe logits:
\begin{equation}
    \hat{p}(y \mid x) = \operatorname{softmax}(g_\phi(f_\theta(x))).
\end{equation}

\paragraph{Temperature Scaling.}
Temperature scaling~\cite{guo2017calibration} fits a single scalar $T > 0$
by minimizing NLL on $\mathcal{D}_\text{cal}$:
\begin{equation}
    T^* = \operatorname*{arg\,min}_{T > 0}
    \;{-\frac{1}{|\mathcal{D}_\text{cal}|}}
    \sum_{(x,y)\in\mathcal{D}_\text{cal}}
    \log \operatorname{softmax}\!\left(\tfrac{g_\phi(f_\theta(x))}{T}\right)_y.
\end{equation}
We parameterize $T = \exp(\log T)$ and optimize with L-BFGS, leaving the
probe's decision boundaries unchanged.

\paragraph{Deep Ensemble.}
We train $M = 5$ independent linear probes, each with a different random seed
and optimized with AdamW on $\mathcal{D}_\text{train}$, with diversity arising
from weight initialization and stochastic gradient noise~\cite{lakshminarayanan2017simple}.
The ensemble prediction averages the member softmax outputs:
\begin{equation}
    \hat{p}_\text{ens}(y \mid x)
    = \frac{1}{M}\sum_{m=1}^{M}
    \operatorname{softmax}\!\left(g_{\phi_m}(f_\theta(x))\right)_y.
\end{equation}

\paragraph{Gaussian-process probe.}
As an exploratory alternative to the linear head, we replace $g_\phi$ with a
sparse variational Gaussian process (SVGP) classifier \cite{hensman2015scalable} on the same frozen
embeddings, using a linear kernel and inducing points fitted on
$\mathcal{D}_\text{train}$. A GP head yields a distribution over functions
rather than a point estimate, so predictive uncertainty is intrinsic to the
model. Unlike the linear probe, the SVGP head did not fit reliably
out of the box: our initial configuration used a Mat\'ern kernel, which gave
unstable, high-variance fits for several encoders, and reaching results we
trust required switching to a linear kernel along with additional tuning of
inducing-point initialization and input dimensionality. With this linear-kernel
configuration we more closely recovered the linear probe's clean accuracy,
confirming the SVGP head was not simply underfitting. We report the
linear-kernel results throughout, for all 16 encoders on all
four classification datasets. We evaluate the SVGP probe under the same corruptions to test
whether a Bayesian head is more robust than previous approaches.

\subsection{Distribution Shift Conditions}
\label{sec:corruptions}

We apply synthetic corruptions at five severity levels to every classification
and segmentation dataset. Classification is stress-tested under all three
corruption types below (cloud/shadow, Poisson--Gaussian sensor noise, motion
blur); segmentation is stress-tested under Poisson--Gaussian sensor noise and
motion blur only. We exclude synthetic cloud from the segmentation sweep by
design: CloudSEN12 already contains real cloud as its labeled target, so a
cloud corruption there would corrupt the label itself, and running a corruption
set that is uniform across all segmentation datasets avoids caveating a single
dataset out of every cloud figure.
The three corruptions are chosen to span distinct physical degradation
mechanisms rather than variations of one: scene-level occlusion (cloud and
shadow), optical smearing from platform motion (motion blur), and detector shot
and read noise (Poisson--Gaussian). Real EO degradation is of course broader,
but sampling three mechanistically unrelated corruptions lets us ask whether a
calibration effect is specific to one corruption or common to all of them; an
effect that appears under every one cannot be an artifact of any single
degradation type.
All corruptions are applied in raw pixel space before normalization
and seeded deterministically by (base seed, image index) for reproducibility
across UQ methods and severity levels.
Corrupting before normalization is likewise deliberate: at deployment only the
training-set statistics are available, so the normalization mismatch a shifted
input induces is itself part of the shift a model sees in practice. These
corruptions are physically motivated but synthetic; they approximate rather than
replace real EO shift (Section~\ref{sec:related_work}). Fig.~\ref{app:corruption_examples}
in the appendix shows example images at each severity level for all three
corruption types.

\paragraph{Cloud and shadow}
Cloud masks are synthesized using the
\texttt{satellite-cloud-generator} library~\cite{rs15174138}, producing
Perlin-noise-based patterns with soft edges and co-registered shadow offsets.
Severity levels 1--5 vary the fraction of pixels under any cloud from
approximately 20\% to full coverage, with opacity tuned per dataset to match sensor-specific
dynamic ranges.
Only optical bands are affected.

\paragraph{Poisson--Gaussian sensor noise}
We model CCD/CMOS detector degradation as~\cite{foi2008practical}:
\begin{equation}
    \tilde{x}_c = \frac{\operatorname{Poisson}(\eta_c \cdot x_c)}{\eta_c}
                  + \sigma_c \cdot \epsilon_c,
    \quad \epsilon_c \sim \mathcal{N}(0, 1),
\end{equation}
where $\eta_c$ controls shot noise intensity and $\sigma_c$ is the readout
noise for channel $c$, set empirically per sensor type (Sentinel-2, Landsat,
aerial).
Severity levels 1--5 progressively reduce $\eta_c$ and increase $\sigma_c$.

\paragraph{Motion blur}
Satellite platform motion during acquisition causes spatial smearing in the
flight direction~\cite{sieberth2014motion,zhu2022blind}.
We simulate this as a horizontal uniform averaging kernel of width
$k \in \{3, 5, 9, 15, 21\}$ pixels for severities 1--5, applied identically
across all channels with reflect padding, following \cite{li2026reobench}.

\subsection{Training-Data Budget Sweep}
\label{sec:budget_protocol}

To gain insights into model behavior under varying training data budgets that are related to promised performance gains of GeoFMs through label efficiency, we retrain each probe at five
training fractions, $\{0.01, 0.10, 0.25, 0.50, 0.75\}$ of the available
training data, on both classification and segmentation. This sweep uses
five random seeds per configuration for classification and three for
segmentation (subset draw plus probe initialization), which lets us separate
genuine rank instability from seed noise.
For each fraction we compute ECE (pixel-ECE for segmentation) and accuracy
(mIoU for segmentation), and measure rank stability by Kendall $\tau$ between
the ranking at each fraction and the full-data ($0.75$) ranking.

\subsection{Selective Prediction}
\label{sec:selective_protocol}

Deployment often permits abstention: a model may defer its least-confident
predictions to an additional processing step such as manual human inspection,
following the selective-classification (reject-option) framework of
\cite{el-yaniv2010foundations, geifman2017selective}. We evaluate whether confidence remains a useful abstention signal under shift using two metrics: selective accuracy at
$90\%$ coverage (accuracy on the $90\%$ most-confident predictions), and the excess
area under the risk--coverage curve (eAURC), which summarizes the quality of the
same confidence ranking without fixing a coverage level.
A model whose confidence adjusts for incorrect predictions can
``defer out'' of its errors; one that is confidently wrong cannot.

\subsection{Evaluation Metrics}
\label{sec:metrics}

We report several metrics that enhance the scope of performance analysis beyond accuracy metrics.

\paragraph{Expected Calibration Error (ECE).}
ECE~\cite{naeini2015ece,guo2017calibration} measures the average gap between
predicted confidence and empirical accuracy across $B = 15$ equal-width bins:
\begin{equation}
    \mathrm{ECE} = \sum_{b=1}^{B}
    \frac{|S_b|}{N}\,\bigl|\operatorname{acc}(S_b)
    - \operatorname{conf}(S_b)\bigr|,
\end{equation}
where $S_b$ is the set of samples whose maximum predicted probability falls
in bin $b$, $\operatorname{acc}(S_b)$ is the fraction correctly classified,
and $\operatorname{conf}(S_b)$ is the mean maximum probability.

\paragraph{Negative Log-Likelihood (NLL).}
NLL is a proper scoring rule~\cite{gneiting2007strictly} sensitive to the full
predictive distribution rather than only the top-class confidence:
\begin{equation}
    \mathrm{NLL} = -\frac{1}{N}\sum_{i=1}^{N} \log \hat{p}(y_i \mid x_i).
\end{equation}

\paragraph{Brier Score.}
The multiclass Brier score~\cite{brier1950verification} is a proper scoring
rule measuring mean squared error between the predicted probability vector and
the one-hot target:
\begin{equation}
    \mathrm{BS} = \frac{1}{N}\sum_{i=1}^{N}\sum_{k=1}^{K}
    \bigl(\hat{p}_k(x_i) - \mathbf{1}[y_i = k]\bigr)^2,
\end{equation}
with range $[0, 2]$.

\paragraph{Signed ECE.}
To distinguish the \emph{direction} of miscalibration we also report the
signed calibration gap $\overline{\operatorname{conf}} -
\overline{\operatorname{acc}}$ (mean confidence minus mean accuracy). Positive
values indicate overconfidence, negative values indicate underconfidence.

\paragraph{Segmentation metrics.}
For dense prediction we compute pixel-ECE (ECE over per-pixel top-label
confidences) and mIoU, mirroring the classification calibration/accuracy pair.

\paragraph{Rank stability (Kendall $\tau$).}
To quantify how much a model ordering changes across conditions we use Kendall's
$\tau$ rank correlation~\cite{kendall1945treatment} between the ranking under one
condition (e.g.\ a corruption severity or data budget) and a reference ranking
(clean, or full-data). Values near $1$ indicate a preserved ordering, near $0$ no
association, and negative values an inverted ordering.

\subsection{Representation Drift Analysis}
\label{sec:cka}

We use linear Centered Kernel Alignment (CKA)~\cite{kornblith2019similarity} to measure how each frozen
encoder's intermediate representations change under corruption.
Intuitively, CKA asks whether the \emph{population-level geometry} of the
representation space survives corruption: it is invariant to rotations and
rescalings of the feature axes, so it can stay high even when individual
points move substantially, as long as their relative arrangement is
preserved.
Given two activation matrices $X, Y \in \mathbb{R}^{N \times D}$ collected
from the same layer under clean and corrupted inputs respectively,
\begin{equation}
    \mathrm{CKA}(X, Y)
    = \frac{\lVert \tilde{Y}^{\top} \tilde{X} \rVert_F^2}
           {\lVert \tilde{X}^{\top} \tilde{X} \rVert_F \cdot \lVert \tilde{Y}^{\top} \tilde{Y} \rVert_F},
    \label{eq:cka}
\end{equation}
where $\tilde{X}$ and $\tilde{Y}$ are the column-centred activation matrices
(each feature zero-meaned across the $N$ samples).
In practice we apply the unbiased HSIC correction of~\cite{nguyen2020wide}.

Forward hooks are placed at four evenly-spaced intermediate layers per encoder.
All encoders are included except RCF, which lacks intermediate semantic
representations.
For each layer and corruption condition we record three quantities. The
first is CKA (Eq.~\ref{eq:cka}) between the paired clean and corrupted
activation matrices, capturing population-level geometry as above. The
second is mean per-sample cosine similarity
$\frac{1}{N}\sum_i \cos(x_i^{\mathrm{clean}},\, x_i^{\mathrm{corr}})$: unlike
CKA, this tracks each sample's \emph{own} displacement directly, so it can
fall even when CKA stays high, i.e., a population whose geometry is
preserved but whose individual points have all shifted. The third is the
participation ratio $(\sum_k \lambda_k)^2\!/\sum_k \lambda_k^2$
(with $\lambda_k$ the squared singular values of the activation matrix),
which measures effective representational dimensionality -- how many
dimensions carry non-trivial variance. A drop in participation ratio under
corruption indicates dimensional collapse, a regime that CKA would not
detect given its insensitivity to low-variance
directions~\cite{ding2021grounding}. The three metrics are therefore
complementary rather than redundant: CKA is a structural, rotation-invariant
comparison; cosine similarity is a metric that directly reflects a
given sample's own movement; and participation ratio catches collapse onto
fewer effective dimensions, which the other two would not flag on their own.
For the clean condition, CKA and cosine similarity are evaluated on a random
split-half of the test activations to establish a noise floor.

At the final layer we additionally examine whether per-sample representation
drift is associated with calibration performance.
For each corrupted condition we compute the Spearman rank correlation between
per-sample drift $\|x_i^{\mathrm{corr}} - x_i^{\mathrm{clean}}\|_2$ and
the probe's maximum softmax confidence, and the same correlation with prediction
correctness.
We also record the fraction of test samples with drift above its median,
confidence above~$0.9$, and prediction wrong: samples that are highly
displaced in representation space yet remain overconfident.
These statistics use the same linear probe as Section~\ref{sec:probe},
trained on clean embeddings and applied to corrupted test inputs without
re-training.

\section{Results}
\label{sec:results}

\subsection{Accuracy Is Not Enough}
\label{sec:clean}

On clean test sets, GeoFMs and general-purpose encoders are indistinguishable
on both accuracy \emph{and} calibration: over the four-dataset classification
canon, GeoFMs average $0.847$ accuracy and $0.034$ ECE and general-purpose
encoders $0.839$ and $0.038$, while the RCF baseline trails on accuracy
($0.739$) without trailing on calibration ($0.039$)
(Table~\ref{app:encoder_severity_full}, group rows).
These ECE values use the same
accuracy-selected regularization strength for every encoder, never re-tuned on
a calibration objective (Section~\ref{sec:probe}), so any miscalibration
reflects the encoder and probe rather than the optimization choice. This clean-data baseline sets up the
question we turn to next: what happens to model performance and rankings when the operating
conditions change (Section~\ref{sec:shift}, Fig.~\ref{fig:tau_severity}).

\subsection{Calibration Rankings Are Unstable Under Shift}
\label{sec:shift}

\begin{figure*}[t]
  \centering
  \includegraphics[width=\textwidth]{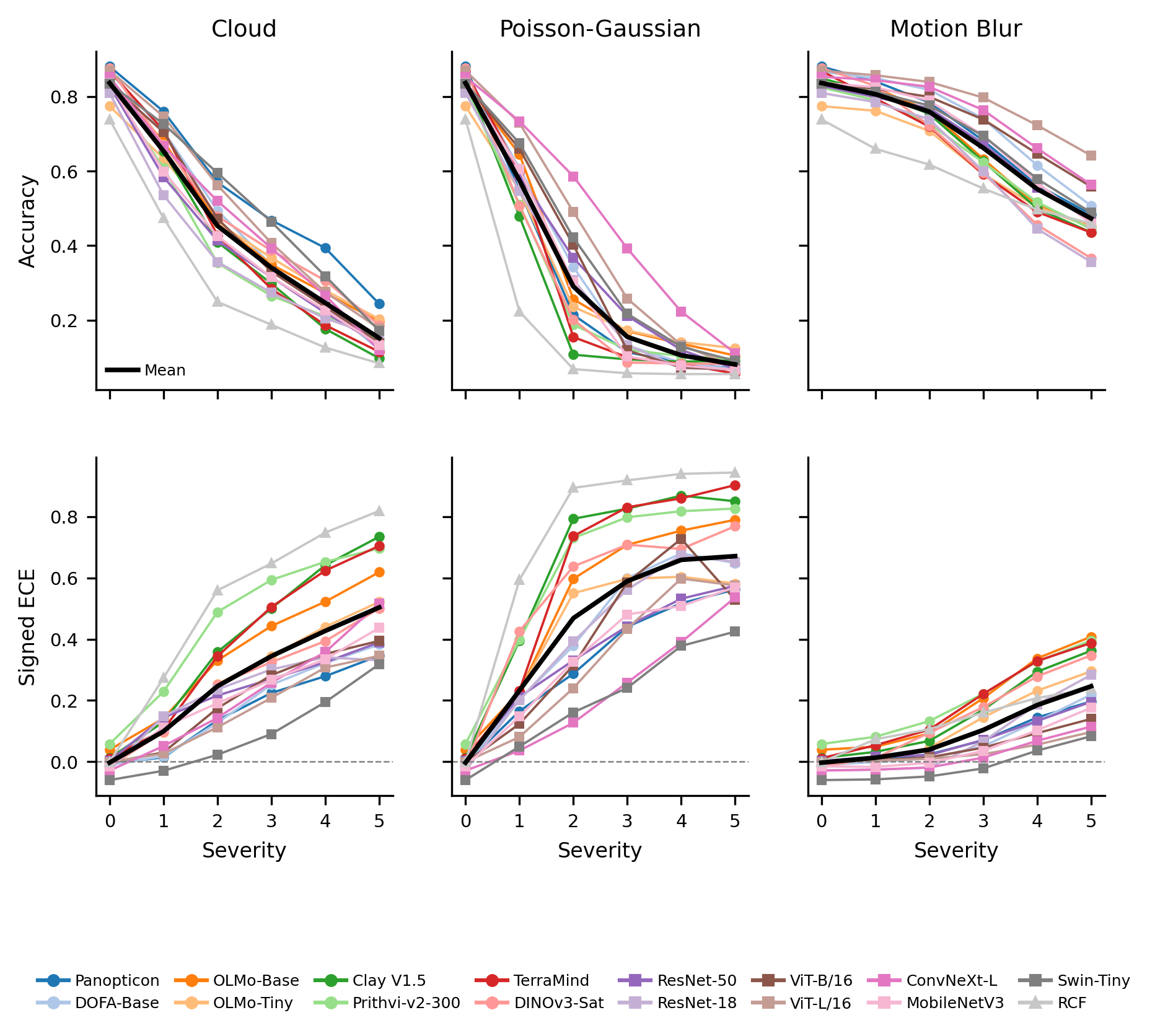}
  \caption{Accuracy (top) and signed ECE (mean confidence $-$ mean accuracy,
  bottom) per model versus severity for three corruption types
  (classification, uncalibrated, mean across the four-dataset canon).
  Accuracy collapses sharply under cloud and Poisson--Gaussian corruption but
  only mildly under motion blur; signed ECE is positive for almost every
  model and grows with severity along the same corruption ordering: failure
  under shift is overconfidence, not underconfidence, and it is worst exactly
  where accuracy is worst. The heavy black line is the mean over the 16
  encoders, which is the quantity quoted in the text. Dashed line (bottom row)
  marks perfect calibration direction.
  (Referenced again in Section~\ref{sec:selective}.)}
  \label{fig:signed_ece}
\end{figure*}

\paragraph{Classification.}
Under physically motivated EO corruptions, calibration degrades sharply:
uncalibrated ECE (mean across the four classification datasets) climbs steeply
with severity (Table~\ref{app:encoder_severity_full}). Under cloud, every encoder
but Swin-Tiny exceeds $0.2$ by severity~3, and at severity~5 the encoders span
$0.32$ to $0.82$; Clay and TerraMind are the worst of the learned encoders
(both about $0.7$), with only the RCF baseline above them. Poisson--Gaussian
sensor noise is more damaging still, pushing TerraMind to $0.83$ and 7 of the 16
encoders past $0.6$ already at severity~3. Motion blur is comparatively mild,
with 13 of 16 encoders still below $0.2$ at severity~3. Which encoder is best
calibrated is itself corruption-dependent: Swin-Tiny, a natural-image supervised
encoder, is the best-calibrated encoder under cloud from severity~2 onward (ECE
$0.09$ at severity~3), but holds no such advantage under motion blur, where it
ranks in the middle of the 16 encoders and ViT-L/16 is clearly better ($0.04$ vs.\ $0.09$ at
severity~3).
The failure is directional. Signed ECE is
positive for almost every model and grows with severity
(Fig.~\ref{fig:signed_ece}), so almost every model becomes \emph{overconfident}.
Accuracy degrades in
step, not merely in proportion to an unchanged error rate: mean top-1
accuracy across the 16 encoders falls from $0.84$ clean to $0.34$ at
cloud severity~3 and $0.15$ at severity~5, and further under
Poisson--Gaussian noise ($0.15$ at severity~3, $0.08$ at severity~5, near
chance); motion blur is comparatively mild, with accuracy still at $0.66$ at
severity~3 and $0.47$ at severity~5 (Fig.~\ref{fig:signed_ece}, top row). The
same corruptions that inflate ECE most severely also collapse accuracy most
severely: calibration and accuracy break down together under cloud and
sensor noise, and degrade only mildly together under motion blur.

\begin{figure*}[t]
  \centering
  \includegraphics[width=\textwidth]{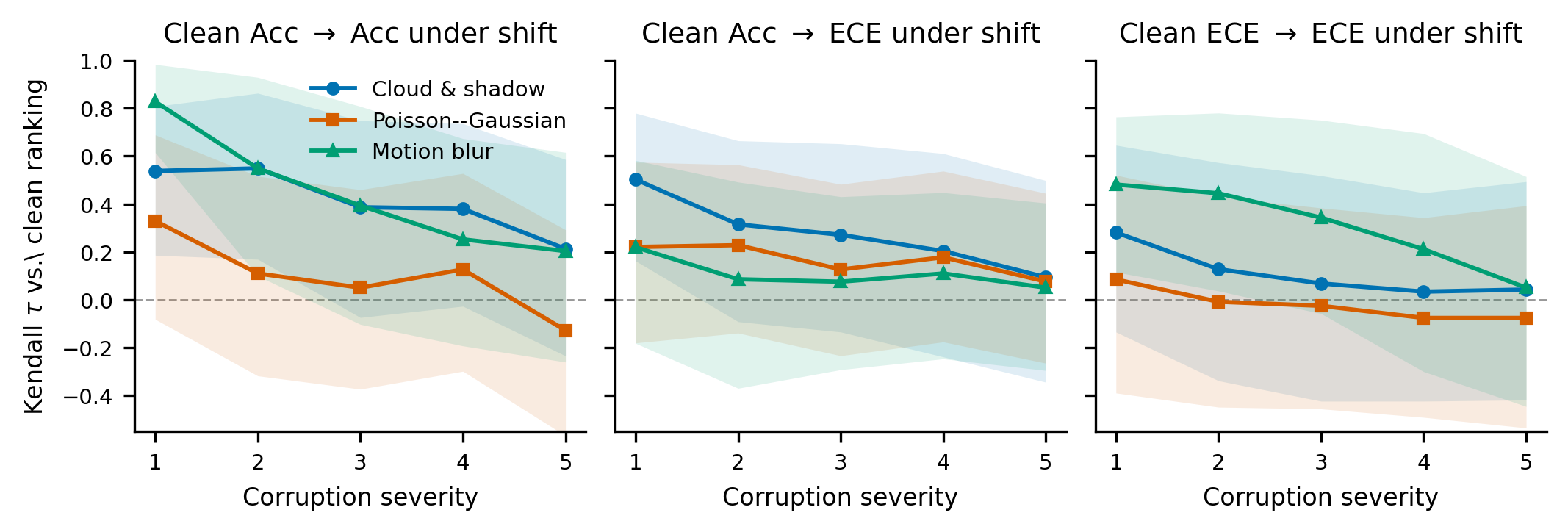}
  \caption{Rank transfer from the clean ranking to the shifted ranking, as a
  function of corruption severity, for all three corruption families
  (classification, 16 encoders, metrics averaged across the four-dataset canon;
  shaded bands are $95\%$ confidence intervals from resampling the
  encoders). Each panel is a different transfer: clean accuracy $\to$ shifted
  accuracy (left), clean accuracy $\to$ shifted ECE (center), and clean ECE
  $\to$ shifted ECE (right). Transfer decays with severity in
  every panel and every family. Accuracy rankings are the most durable, but
  only under the milder corruptions: cloud and motion blur hold $\tau\approx0.4$
  into mid severity, whereas Poisson--Gaussian accuracy transfer falls to
  $\tau=0.05$ by severity~3 and inverts by severity~5 ($\tau=-0.13$). Clean
  rankings say little about shifted \emph{calibration} in any family: the clean
  accuracy ordering agrees with the shifted ECE ordering at $\tau\le0.3$ across
  nearly all cells, and the clean and shifted ECE orderings agree at $\tau=0.07$
  (cloud severity~3) and below.
  The bands are wide at $n=16$: a single $\tau$ cannot be pinned down, so the
  robust signal is the decay itself, which reproduces independently in every
  corruption, rather than any one value.}
  \label{fig:tau_severity}
\end{figure*}

Fig.~\ref{fig:tau_severity} makes this concrete. Each $\tau$ compares two
orderings of the same 16 encoders: with one ranking by their clean-data value, and a second ranking  by their value under corruption, we measure how much of the ordering
survives. Under cloud shift the accuracy ordering is moderately preserved
($\tau=0.39$ from clean to cloud severity~3), whereas the calibration ordering
is not: the clean and shifted ECE orderings agree at $\tau=0.07$ (severity~3)
and $\tau=0.04$ (severity~5), and ranking by clean \emph{accuracy} recovers the
shifted ECE ordering no better ($\tau=0.27$). The decay reproduces independently
in all three corruption families, so the conclusion does not rest on any single
imprecise $\tau$. Once \emph{already} shifted, ECE rankings do cohere among
themselves (ECE severity-3 $\to$ severity-5 $\tau=0.72$, computed from the same
rankings but not plotted in Fig.~\ref{fig:tau_severity}), so the shifted
rankings are structured rather than noise, but knowing which encoder is
best-calibrated on clean data still says little about which is best-calibrated
under the shift a model could actually encounter. The scatter views in the
appendix (Fig.~\ref{app:rank_sensitivity}) show the same instability model by
model.

\begin{figure*}[t]
  \centering
  \begin{minipage}{0.49\textwidth}
    \centering
    \includegraphics[width=\textwidth]{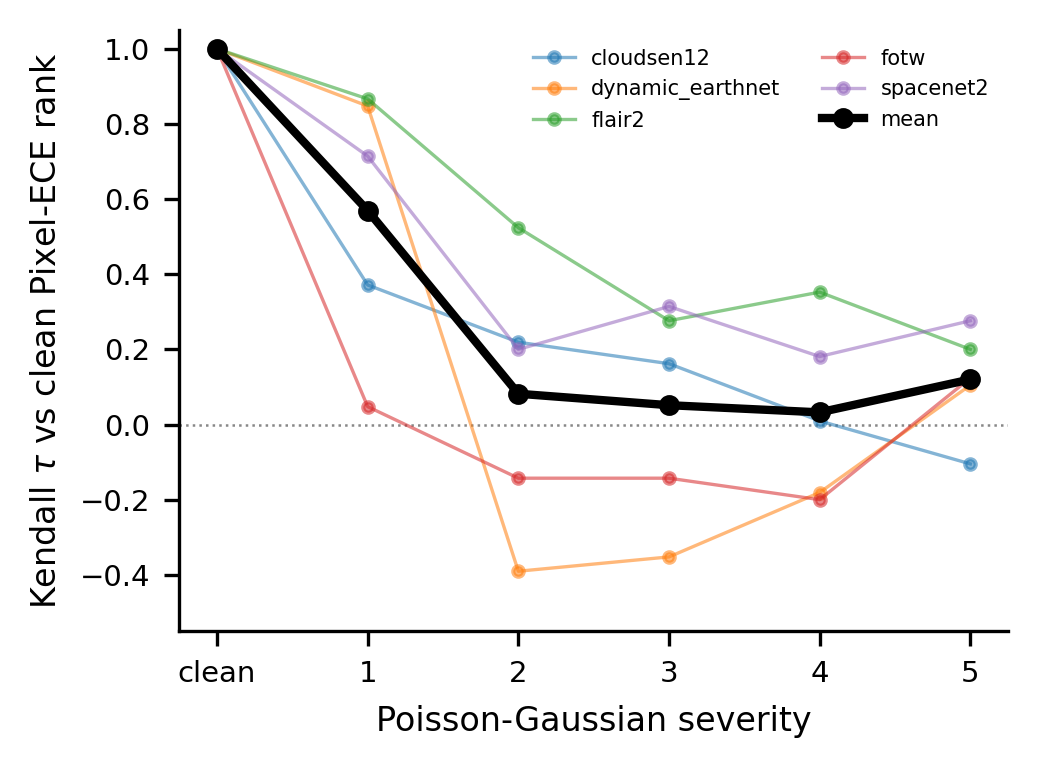}
  \end{minipage}\hfill
  \begin{minipage}{0.49\textwidth}
    \centering
    \includegraphics[width=\textwidth]{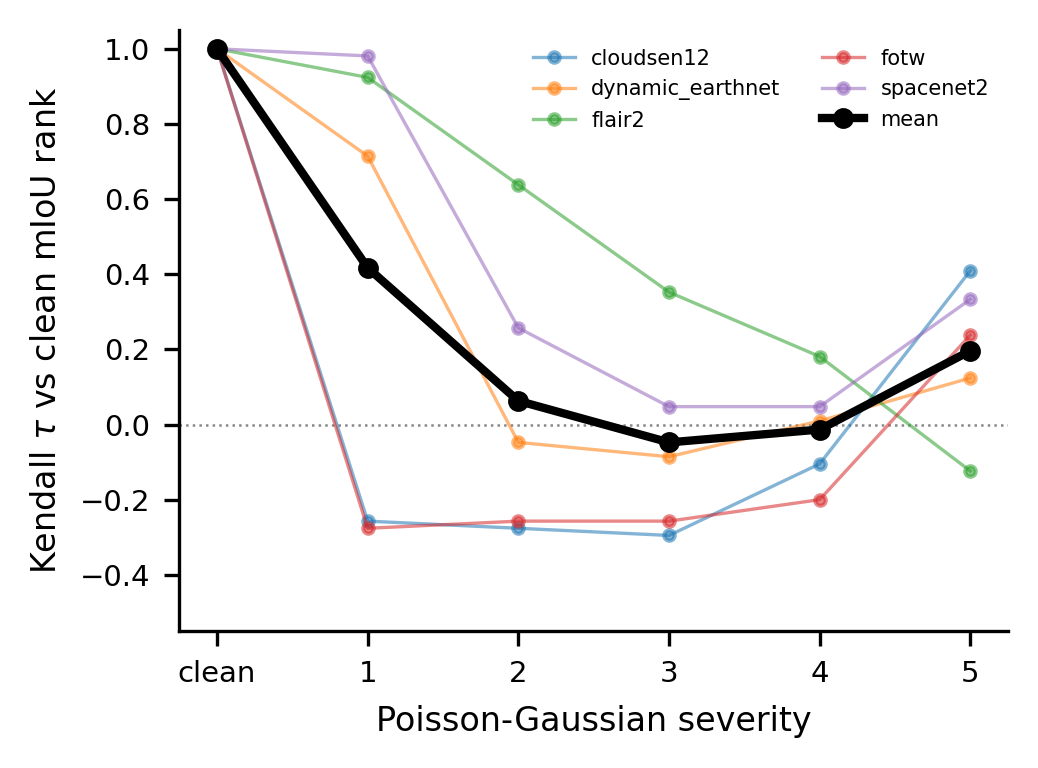}
  \end{minipage}
  \caption{Under Poisson--Gaussian sensor noise, both mIoU and pixel-ECE rankings collapse
  for dense prediction.
  \emph{Left:} pixel-ECE rank stability (Kendall $\tau$ vs.\ the clean
  ranking) collapses from $\approx0.57$ at severity~1 to $\approx0.05$ by severity~3 under
  Poisson--Gaussian noise, aggregated across five segmentation datasets and 15 encoders;
  several datasets invert (negative $\tau$). \emph{Right:} mIoU rank stability under the
  same corruption collapses comparably, from $\approx0.42$ at severity~1 to between
  $\approx-0.05$ and $\approx0.20$ across severities~2--5. The more stable accuracy
  rankings in Fig.~\ref{fig:tau_severity} are measured under \emph{cloud} shift; matched on
  corruption, classification accuracy rankings transfer no better than these
  ($\tau=0.05$ at severity~3 and $-0.13$ at severity~5 under Poisson--Gaussian noise), so
  the accuracy-vs-calibration asymmetry is specific to the corruption rather than a
  property of dense prediction.}
  \label{fig:seg_corruption}
\end{figure*}

\paragraph{Segmentation.}
The pixel-ECE instability is not a classification artifact: for segmentation, the
pixel-ECE ranking against clean collapses from $\tau\approx0.57$ at severity~1
to near zero (with per-dataset inversions) by severity~2 and beyond. Here, however,
the mIoU ranking collapses nearly as much over the same severities
($\tau\approx0.42\to-0.05$--$0.20$; Fig.~\ref{fig:seg_corruption}). The asymmetry we saw
under cloud shift and in the training-budget comparisons (\S\ref{sec:budget}) is absent
here, but this tracks the corruption rather than the task: classification accuracy
rankings transfer just as poorly under Poisson--Gaussian noise ($\tau=0.05$ at severity~3,
$-0.13$ at severity~5). Under this corruption, neither accuracy nor calibration rankings
survive.

\begin{figure}[t]
  \centering
  \includegraphics[width=\columnwidth]{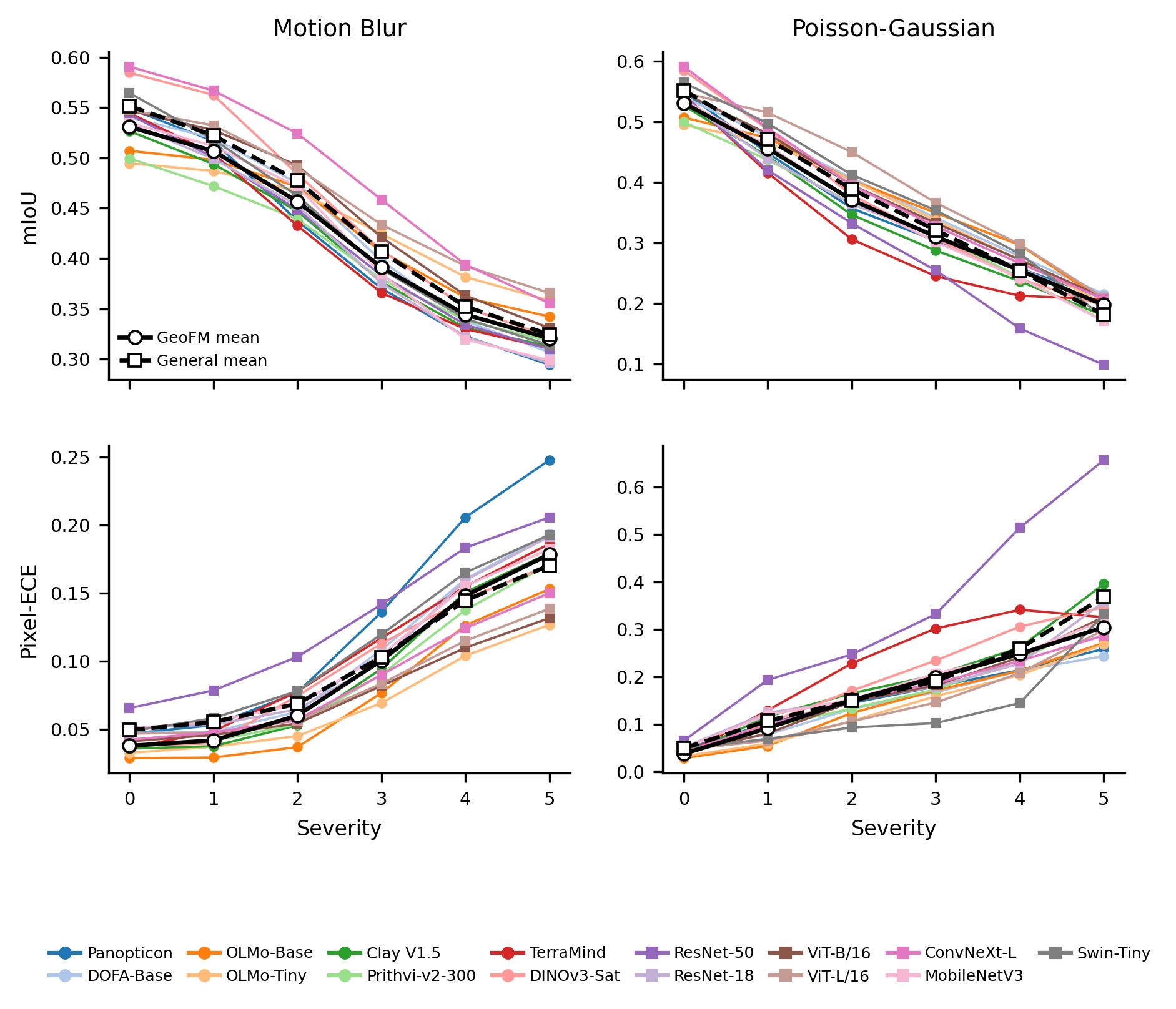}
  \caption{Per-model mIoU (top) and pixel-ECE (bottom) versus severity for
  segmentation (15 encoders, no RCF, mean across five datasets); marker shape
  denotes domain (circle GeoFM, square general-purpose); the two heavy black
  lines are the domain means, solid for GeoFM and dashed for general-purpose.
  The ratios quoted below average each encoder's own severity-5/clean ratio, so
  they are close to but not identical to the quotient of the two plotted domain
  means. mIoU degrades
  comparably across domains, but the pixel-ECE degradation is
  domain-conditioned on average: GeoFM encoders reach a larger sev-5/clean ECE
  ratio than general-purpose encoders under both corruptions (motion blur
  $4.75\times$, std $0.56$, vs.\ $3.47\times$, std $0.37$; Poisson--Gaussian
  $8.30\times$, std $2.00$, vs.\ $7.34\times$, std $1.30$), though individual
  encoders vary. Under motion blur the two largest ratios are OLMoEarth-base
  ($5.34\times$) and Panopticon ($5.31\times$), both EO-pretrained but with
  different pretraining objectives; under Poisson--Gaussian noise the
  general-purpose ResNet-50 ($10.03\times$) is second only to the GeoFM Clay
  ($11.08\times$).}
  \label{fig:seg_by_model}
\end{figure}

Finally, the ECE degradation is domain-conditioned rather than objective-conditioned:
GeoFM encoders show a larger sev-5/clean ECE ratio than general-purpose encoders
under both corruptions (Fig.~\ref{fig:seg_by_model}), even though mIoU degrades
comparably across domains. A training-objective-level breakdown (EO-recon/EO-distill/Nat-DINO/
Nat-supervised) is noisier because group
sizes are unbalanced (EO-distill $n=2$, Nat-DINO $n=1$), and Panopticon (EO-distill)
is among the worst degraders under motion blur ($5.31\times$, essentially tied
with OLMoEarth-base at $5.34\times$; Fig.~\ref{fig:seg_by_model} caption). This motivates the mechanism in Section~\ref{sec:mechanism}.

\subsection{A Candidate Representational Mechanism, Conditioned on Pretraining Domain}
\label{sec:mechanism}

\begin{figure*}[t]
  \centering
  \includegraphics[width=\textwidth]{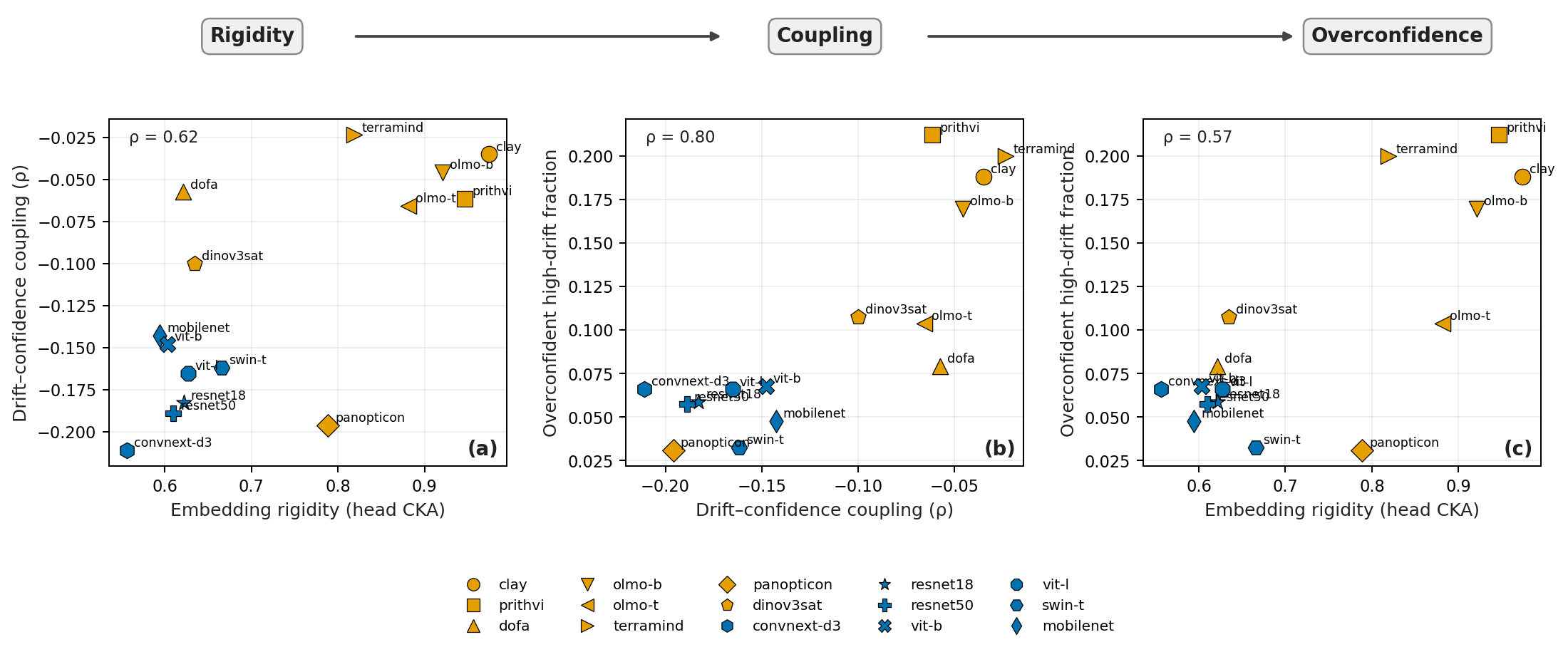}
  \caption{\textbf{Rigidity, decoupling, and overconfidence move together
  across encoders.} Each point
  is one encoder averaged over the four-dataset classification canon and three
  corruptions; color is pretraining domain (orange EO, blue natural-image).
  \textbf{(a)} More rigid embeddings
  (higher head-layer CKA under corruption) tend to have weaker drift--confidence
  coupling ($\rho=0.62$). \textbf{(b)} Weaker coupling goes with a larger fraction of
  samples that are simultaneously high-drift, high-confidence, and wrong
  ($\rho=0.80$). \textbf{(c)} More rigid embeddings likewise go with more
  overconfident high-drift errors ($\rho=0.57$). The three panels share the same
  encoders, so they are not independent tests. The natural-image encoders cluster
  tightly (CKA $0.56$--$0.67$) while the EO-pretrained encoders spread across
  $0.62$--$0.97$: Clay, Prithvi, OLMo, and TerraMind occupy the high-rigidity,
  low-coupling, high-overconfidence region, whereas Panopticon is rigid ($0.79$)
  yet the least overconfident encoder we evaluate ($0.03$), showing rigidity
  is not sufficient for overconfidence.}
  \label{fig:track_b_chain}
\end{figure*}

To see how pretraining domain relates to calibration collapse, we report four
correlational measurements on the four-dataset classification canon
(Fig.~\ref{fig:track_b_chain}), computed at a single seed. Each links EO
pretraining to overconfident errors under corruption. They agree with one
another, but they are correlational, so we treat them as evidence for a
hypothesis rather than a proven cause, and state that hypothesis at the end of
the section.

\textbf{(B1) Rigidity.} EO-pretrained encoders barely move under corruption: their
head-layer CKA between clean and corrupted inputs stays highest
($\approx0.82$), well above the natural-image
encoders ($\approx0.61$; $p=0.004$, two-sided Mann--Whitney $U$ test~\cite{mann1947test}, a
nonparametric two-sample test suited to the small, non-normal per-encoder CKA samples, $n=8$ vs.\ $7$;
Fig.~\ref{fig:track_b_dotplot} in the
appendix, rigidity panel). A plausible explanation is that
EO pretraining rewards perturbation-invariant features, producing embeddings
whose variance concentrates in far fewer directions on clean data
(participation ratio $\approx13$, versus $\approx31$ for the natural-image
encoders; Fig.~\ref{app:participation_ratio}) and which therefore change little
when the input is degraded.
\textbf{(B2) Decoupling.} For EO-pretrained encoders, confidence changes little
as the input drifts: drift--confidence coupling is near zero ($\rho\approx-0.07$)
versus clearly negative for natural-image encoders ($\rho\approx-0.17$; $p=0.009$;
Fig.~\ref{fig:track_b_dotplot}, coupling panel). This is consistent with rigid
embeddings holding the probe's logits, and hence its confidence, roughly fixed
under corruption.
\textbf{(B3) Overconfident errors.} EO-pretrained encoders also show a
$\sim\!2.4\times$ higher fraction of high-drift, high-confidence, wrong predictions
($\approx0.14$ vs.\ $\approx0.06$; $p=0.014$;
Fig.~\ref{fig:track_b_dotplot}, overconfidence panel).

\begin{figure*}[t]
  \centering
  \includegraphics[width=\textwidth]{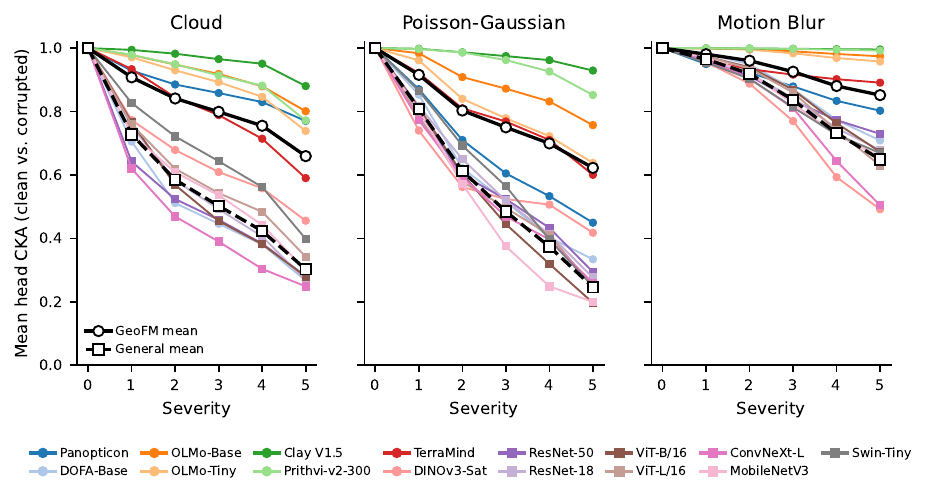}
  \caption{Head-layer CKA(clean, corrupted) versus severity, one line per
  encoder (classification, four-dataset canon); the two heavy black lines are
  the domain means within each panel, solid for EO-pretrained and dashed for
  natural-image. Similarity declines for every
  encoder as severity increases, but at a domain-conditioned rate: EO-pretrained
  encoders stay more rigid at every severity (mean CKA $0.93\to0.71$ from
  severity~1 to~5, averaging the three panels) than natural-image encoders
  ($0.83\to0.40$). The split is not
  uniform: Clay and OLMo-base remain above CKA~$0.8$ even at severity~5 under
  cloud and motion blur, while
  DOFA drops into the natural-image range ($0.44$) and TerraMind falls to an
  intermediate $0.69$.}
  \label{fig:cka_similarity_severity}
\end{figure*}

\begin{table}[t]
  \centering
  \caption{Cross-severity robustness of the CKA--ECE association. Spearman
  $\rho$ between head-layer CKA(clean, corrupted) and ECE inflation across the
  15 encoders that admit a layerwise CKA, per corruption family
  and severity; $p$ from the analytic Spearman test. Every cell is positive
  (median $\rho=0.39$); 6 of 15 reach $p<0.05$. The RCF baseline is excluded
  because it has only two layers rather than the depth-sampled blocks used for
  the learned encoders, so a layerwise CKA is not comparable for it.}
  \label{tab:cka_ece_robustness}
  \begin{tabular}{lcccccc}
    \toprule
     & \multicolumn{2}{c}{Cloud} & \multicolumn{2}{c}{Motion blur}
       & \multicolumn{2}{c}{Poisson--Gauss.} \\
    \cmidrule(lr){2-3}\cmidrule(lr){4-5}\cmidrule(lr){6-7}
    Sev. & $\rho$ & $p$ & $\rho$ & $p$ & $\rho$ & $p$ \\
    \midrule
    1 & 0.31 & .260 & 0.24 & .390 & 0.27 & .334 \\
    2 & 0.59 & .021 & 0.33 & .226 & 0.37 & .168 \\
    3 & 0.52 & .044 & 0.56 & .030 & 0.34 & .221 \\
    4 & 0.50 & .056 & 0.60 & .017 & 0.39 & .156 \\
    5 & 0.55 & .035 & 0.52 & .044 & 0.36 & .187 \\
    \bottomrule
  \end{tabular}
\end{table}

Rigidity is not confined to the single severity used above. Tracking head-layer
CKA across the full severity range (Fig.~\ref{fig:cka_similarity_severity})
shows every encoder's representation does move as corruption intensifies, so
rigid encoders are not frozen at a fixed value; what differs is the rate.
EO-pretrained encoders decline from mean CKA $0.93$ to $0.71$ between
severities~1 and~5, natural-image encoders from $0.83$ to $0.40$, the same
domain split as (B1).

Head-layer CKA also tracks calibration failure directly, not only through the
composite overconfident-error fraction of (B3). For each of 15 conditions (three
corruption families times five severities) we rank the 15 encoders by mean
head-layer CKA and by ECE inflation, and correlate the two orderings
(Table~\ref{tab:cka_ece_robustness}). Every condition gives a positive
correlation (median $\rho=0.39$) and 6 reach $p<0.05$: cloud at severities~2, 3
and~5, motion blur at~3, 4 and~5. The association strengthens with severity
under cloud and motion blur and is weaker, though still positive, under
Poisson--Gaussian noise. The conditions share the same encoders, so they are not
independent tests and no single one is decisive; cloud severity~3, shown per
encoder in Fig.~\ref{fig:cka_similarity_ece}, is a typical cell ($\rho=0.52$,
$p=.044$).

Taken together, (B1) through (B3) and the CKA--ECE robustness are four
correlational measurements over our selection of models, so they motivate rather than
establish a mechanism: if EO pretraining holds the representation still under
corruption, the probe never sees the movement it would need to lower its
confidence, and the errors it makes stay confident. However, we caveat it with two points. First, Panopticon is rigid
yet the least overconfident encoder we evaluate, so rigidity is associated with,
but not sufficient for, overconfidence; its any-sensor pretraining is a candidate
explanation that we leave untested, and with a single any-sensor encoder we
cannot generalize it. Second, confirming that rigidity causes calibration failure
would require dedicated controlled experiments, for instance encoders matched on
everything but pretraining domain, which we leave to future work.

\subsection{The Second Axis: Training-Data Budget}
\label{sec:budget}

Distribution shift is one stress axis; the amount of training data is the
other, and it is the one very commonly associated with the promise of label efficiency on diverse downstream tasks for GeoFMs \cite{zhu2026foundations}.

\paragraph{Classification.}
The pattern from Section~\ref{sec:shift} repeats on this orthogonal axis:
shrinking the training fraction from $75\%$ to $1\%$ leaves accuracy
rankings comparatively stable while calibration rankings collapse. The
small-data regime is where calibration is \emph{worst}.

\begin{figure*}[t]
  \centering
  \begin{minipage}{0.49\textwidth}
    \centering
    \includegraphics[width=\textwidth]{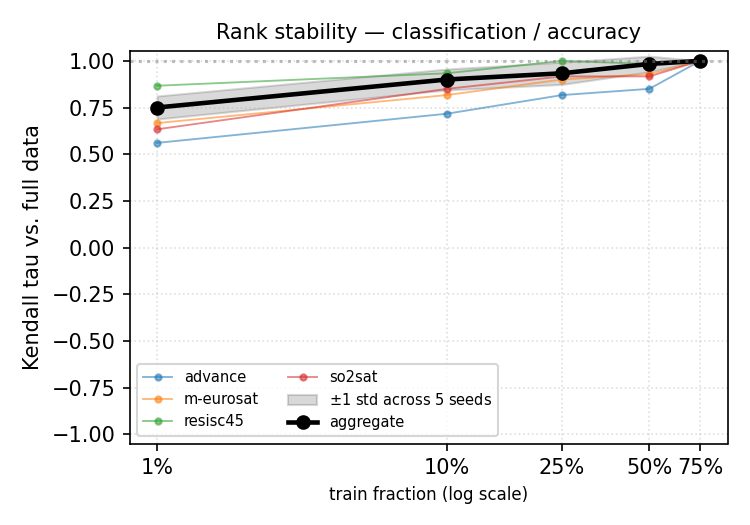}
  \end{minipage}\hfill
  \begin{minipage}{0.49\textwidth}
    \centering
    \includegraphics[width=\textwidth]{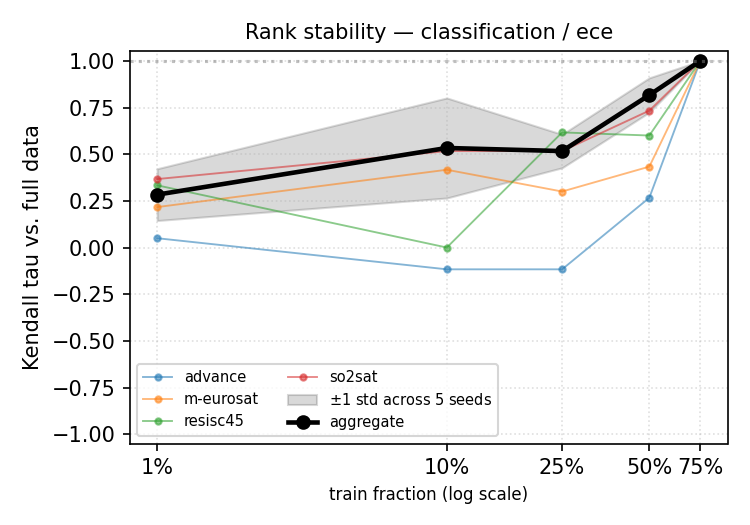}
  \end{minipage}
  \caption{Rank-stability Kendall $\tau$ against the full-data ($75\%$) ranking as the
  training fraction shrinks (classification, mean across the four-dataset canon; shaded
  band is $\pm 1$ std of the aggregate $\tau$ across the five seeds). \emph{Left:} accuracy
  rankings stay comparatively stable, $\tau=0.75$ at $1\%$. \emph{Right:} calibration
  rankings collapse, $\tau=0.28$ at $1\%$, a drop that exceeds the seed-noise band. The
  same contrast as under shift, on an orthogonal axis.}
  \label{fig:tau_budget}
\end{figure*}

Concretely, accuracy rankings are fairly stable across budgets (Kendall
$\tau=0.75$ between the $1\%$ and $75\%$ rankings), while calibration rankings
collapse ($\tau=0.28$ for the same comparison), a drop that visibly exceeds the
five-seed noise band (Fig.~\ref{fig:tau_budget}).

\begin{figure*}[t]
  \centering
  \begin{minipage}{0.49\textwidth}
    \centering
    \includegraphics[width=\textwidth]{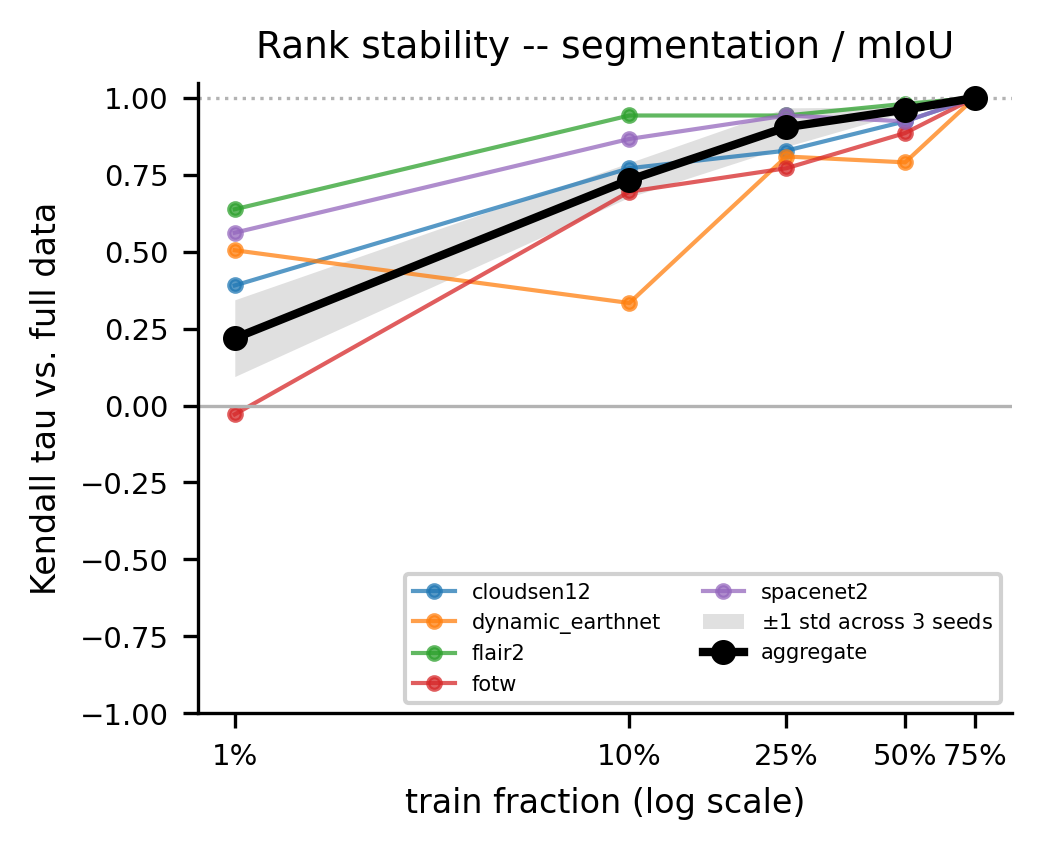}
  \end{minipage}\hfill
  \begin{minipage}{0.49\textwidth}
    \centering
    \includegraphics[width=\textwidth]{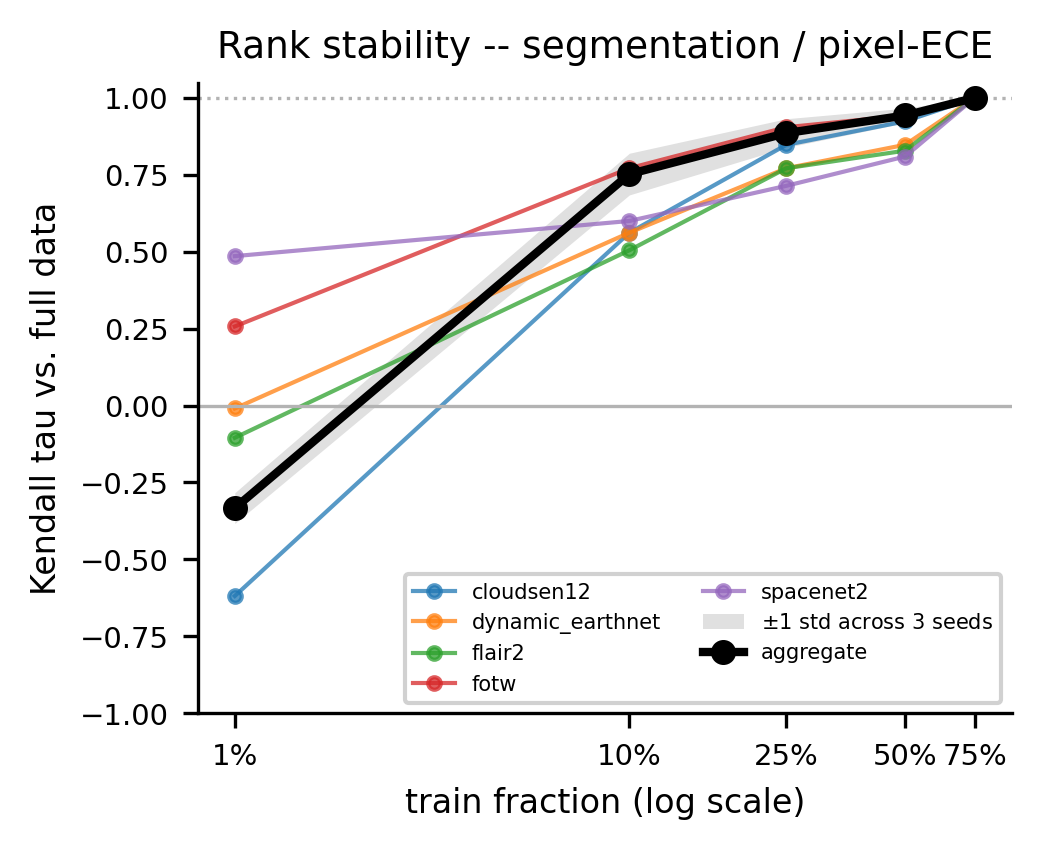}
  \end{minipage}
  \caption{Segmentation training-budget rank-stability Kendall $\tau$ against the full-data
  ($75\%$) ranking (five datasets: CloudSEN12, DynamicEarthNet, FLAIR\#2, Fields of the
  World, SpaceNet2; 15 encoders; three seeds; shaded band is $\pm 1$ std of the aggregate
  $\tau$ across seeds). \emph{Left:} mIoU rankings stay comparatively stable, $\tau=0.22$ at
  $1\%$. \emph{Right:} pixel-ECE rankings collapse and invert ($\tau=-0.33$ at $1\%$), mirroring
  the classification result.}
  \label{fig:tau_budget_seg}
\end{figure*}

\paragraph{Segmentation.}
Segmentation behaves the same way: mIoU rankings
stay comparatively stable ($\tau\approx0.22$ and up) while pixel-ECE rankings invert
($\tau\approx-0.33$; Fig.~\ref{fig:tau_budget_seg}), a gap that again exceeds the
(three-seed) noise band, as for classification.

\subsection{Mitigations Do Not Restore Calibration Under Shift}
\label{sec:mitigations}

This section and Section~\ref{sec:selective} apply to classification datasets only.
If calibration rankings are unstable under deterministic softmax predictions, can a targeted calibration \emph{method} restore them? We test three commonly used UQ methods, but find that none fully solves the encountered issues.

\begin{figure*}[t]
  \centering
  \includegraphics[width=\textwidth]{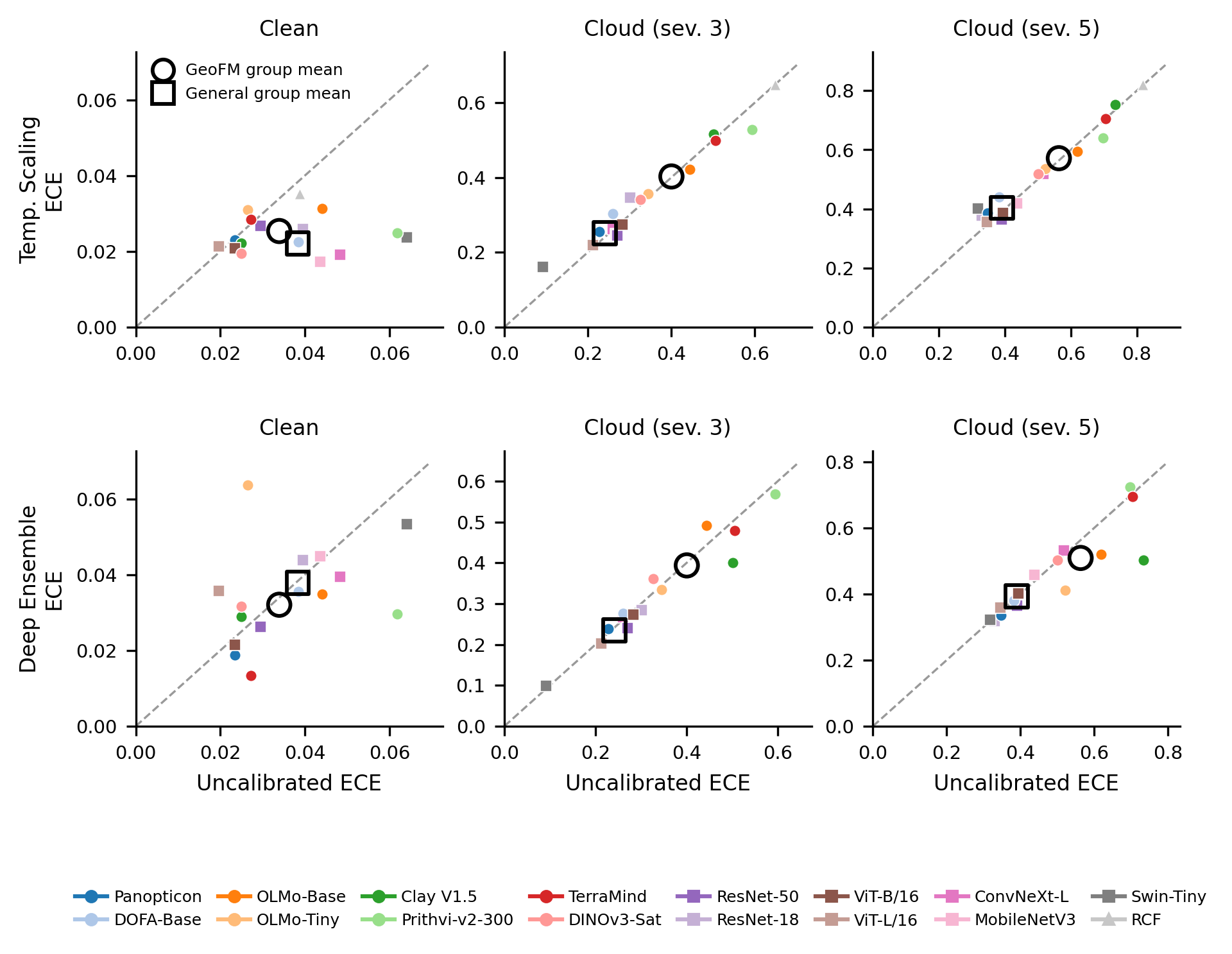}
  \caption{Temperature scaling (top) and deep ensembles (bottom): method ECE
  vs.\ uncalibrated ECE at clean, cloud severity~3, and severity~5 (four-dataset
  canon, mean across datasets, one point per encoder per severity). The two
  large hollow black markers in each panel are the group means, circle for
  EO-pretrained and square for general-purpose. Points
  below the diagonal improve calibration. TS moves points below the diagonal
  on clean data but sits \emph{on} it under shift, consistent with the
  shift-miscalibration failure reported by \cite{ovadia2019can}: at cloud
  severity~3 the EO-pretrained group mean goes from $0.400$ uncalibrated to
  $0.403$ after temperature scaling, and the general-purpose group mean from
  $0.241$ to $0.251$.
  Deep ensembles pull points partially below the diagonal at severity~3 but at
  $k\times$ cost.}
  \label{fig:ts_scatter}
\end{figure*}

\textbf{Temperature scaling} works in-distribution but is fundamentally
constrained by its own protocol: $T^*$ is fit once on clean calibration data
and never re-exposed to corrupted inputs (Section~\ref{sec:probe}), so it
has no way to correct for a shift it was never given information about.
Under this deployment constraint (clean data is what is
available at calibration time), shifted ECE is left essentially unchanged
(Fig.~\ref{fig:ts_scatter}, top; EO-pretrained cloud severity-3 ECE
$0.400\to0.403$), the same clean-only-calibration failure mode reported
by~\cite{ovadia2019can}.
\textbf{Deep ensembles} help partially at cloud severity~3 but require $k$
independent probes and still leave large residual miscalibration
(Fig.~\ref{fig:ts_scatter}, bottom). The help is also uneven across encoders and
does not always hold as severity rises: for Prithvi-v2-300 the ensemble
$\Delta$ECE moves from $-0.009$ at severity~1 and $-0.026$ at severity~3 to
$+0.027$ at severity~5, so the ensemble stops helping and starts hurting
(Table~\ref{app:posthoc_full_fig}). Brier confirms these trends across cloud
severities, but NLL does not track temperature scaling's reversal: its
$\Delta$NLL stays negative through severity~5, where its $\Delta$ECE has already
turned positive by severity~4 (Table~\ref{app:posthoc_full_fig}).

\begin{figure*}[t]
  \centering
  \includegraphics[width=\textwidth]{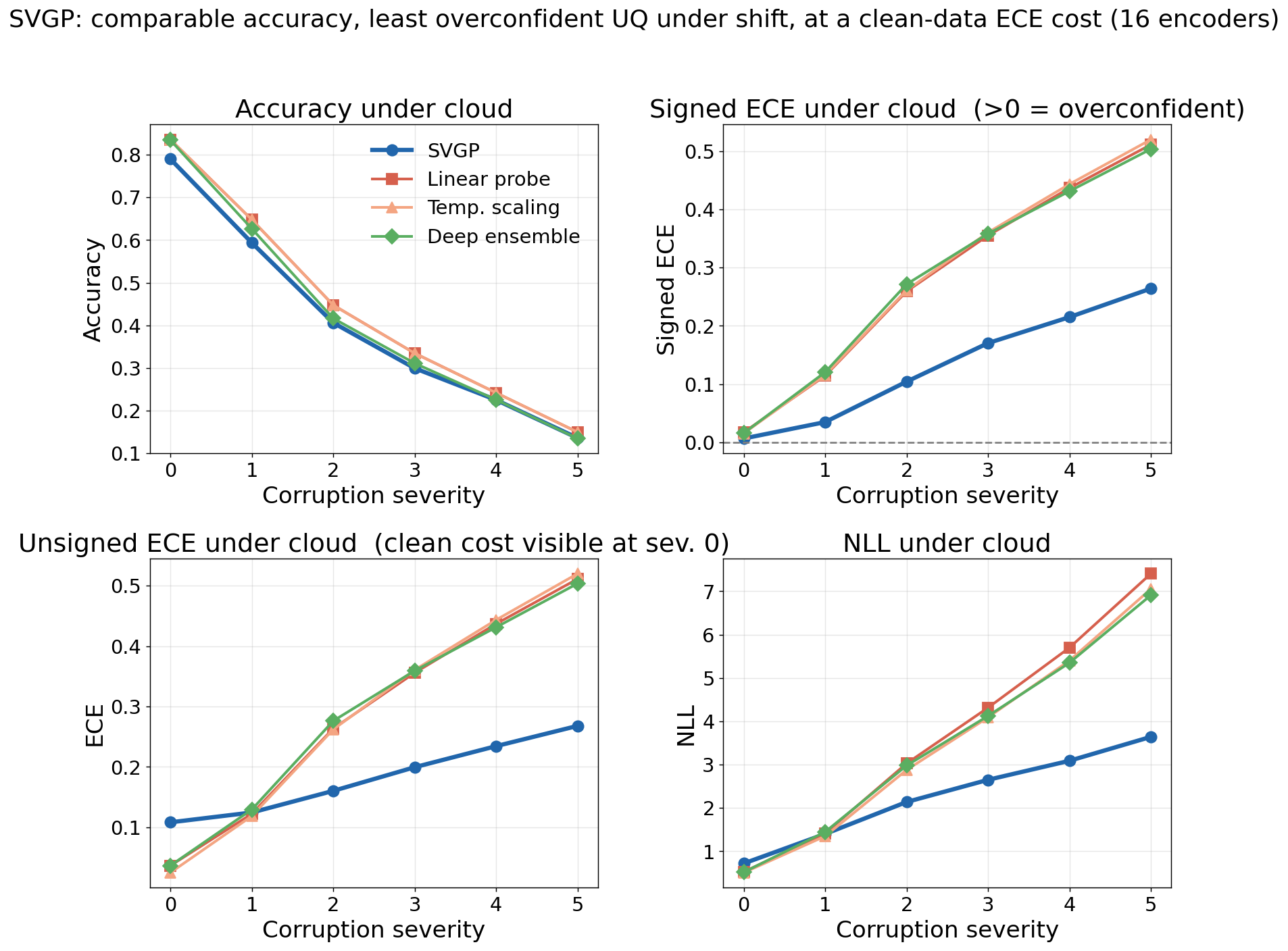}
  \caption{Gaussian-process (SVGP) probe versus linear-probe baselines under
  cloud corruption (classification, mean across the four-dataset canon,
  all 16 encoders). The SVGP head is the least
  overconfident under severe shift (roughly halving signed and unsigned ECE at
  high severity) at comparable accuracy, but at higher training cost and a
  higher clean-data ECE (bottom-left panel: $\approx0.108$ vs.\ $\approx0.036$
  for the linear probe at severity~0).}
  \label{fig:gp}
\end{figure*}

\textbf{A Gaussian-process probe} behaves differently. The SVGP head
is notably less overconfident under severe shift, roughly halving signed ECE
at high cloud severity at comparable accuracy (Fig.~\ref{fig:gp}), but it
raises clean-data ECE by $\sim\!3.0\times$ ($\approx0.108$ vs.\ $\approx0.036$)
and costs more to train. We present it as an exploratory direction. Across all
three methods, the result is consistent: calibration under shift must be
measured and designed for, and is not really possible to fix in a post-hoc manner through these approaches.

\subsection{Selective Prediction Cannot Defer Around Confident Errors}
\label{sec:selective}

The UQ methods above aim to improve the predictive distribution, and with it
the uncertainty estimates a deployment would defer on. Selective prediction
tests one such use directly: whether a model's confidence still separates
correct from incorrect predictions well enough to defer its errors under shift
(Section~\ref{sec:selective_protocol}). The predictions at issue are the
high-drift, high-confidence errors of (B3): $9.9\%$ of test samples averaged
over the 15 encoders that admit the drift analysis, split $0.14$ for the
EO-pretrained encoders against $0.06$ for the natural-image ones
(Fig.~\ref{fig:track_b_dotplot}, overconfidence panel). That is a modest share
in absolute terms, but it is exactly the share abstention should ideally detect.

\begin{figure*}[t]
  \centering
  \includegraphics[width=\textwidth]{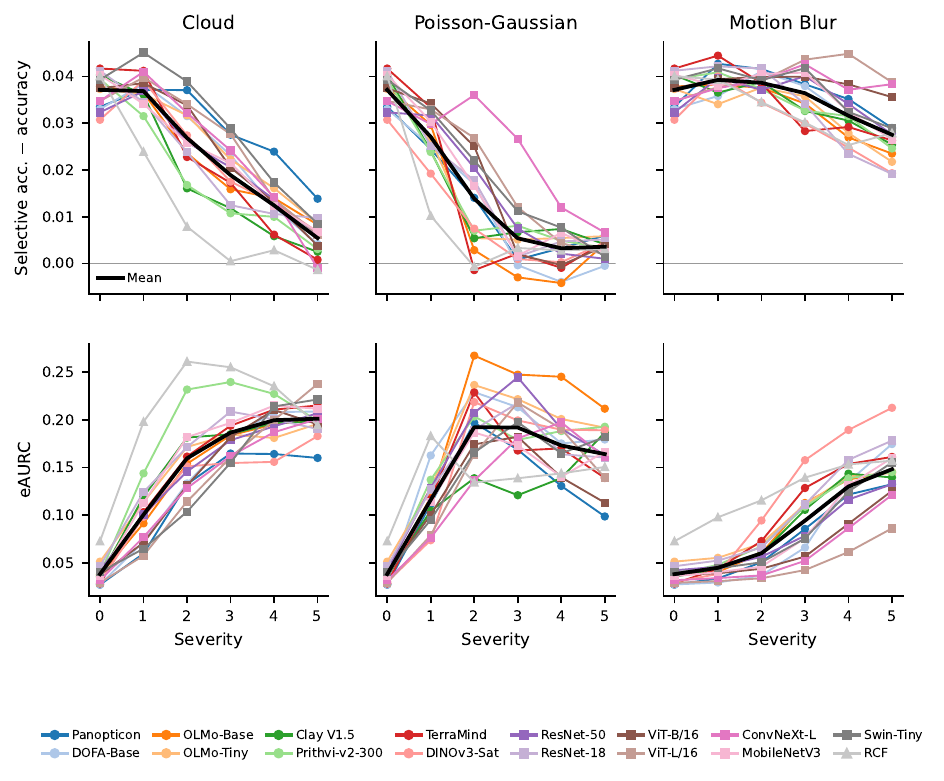}
  \caption{Selective-prediction gain (selective accuracy at $90\%$ coverage
  $-$ overall accuracy, top) and eAURC (bottom) per model versus severity
  (classification, mean across the four-dataset canon); the heavy black line is
  the mean over the 16 encoders, which is the quantity quoted in the text.
  The gain decays toward
  zero, and slightly below it for RCF under cloud and for OLMo-Base, TerraMind
  and DOFA-Base under Poisson--Gaussian noise, as severity rises, while eAURC
  rises: abstention of predictions does not help, because the models are
  confidently wrong. The Poisson--Gaussian eAURC turns down again beyond
  severity~2, where mean accuracy has fallen to $0.29$ and below
  (Fig.~\ref{fig:signed_ece}); with
  accuracy approaching chance there is little risk left for a confidence
  ranking to reduce, so the turn-down is an artifact of the collapse rather
  than a recovery of the confidence signal.}
  \label{fig:selective}
\end{figure*}

\textbf{Selective prediction} is a common deployment
strategy~\cite{el-yaniv2010foundations, geifman2017selective}: if a
model's confidence reflects its accuracy, low-confidence predictions can be
deferred. However, we find that this expectation fails under our employed
distribution shifts. The selective-accuracy gain from
abstaining on the least-confident $10\%$ decays to zero, and turns negative
for the most overconfident encoders, as severity rises
(Fig.~\ref{fig:selective}, top). eAURC moves the same way: from $0.038$ on clean
data it rises to $0.187$ at cloud severity~3 and $0.201$ at severity~5, and from
$0.045$ to $0.148$ across the milder motion-blur range
(Fig.~\ref{fig:selective}, bottom). Because
the errors are confident, deferral cannot recover them: confidence remains
correlated with correctness on clean data, but fails to be a reliable deferral
signal under our employed distribution shifts. In our experiments deployment-time shift therefore
cannot be detected from the model's own confidence, but discuss alternatives in
Section~\ref{sec:discussion}.

\section{Discussion}
\label{sec:discussion}

\paragraph{What the rankings mean for evaluation.}
The recurring pattern across two orthogonal stress axes, distribution shift
(Section~\ref{sec:shift}) and training-data budget (Section~\ref{sec:budget}), is
that model rankings are broadly unstable once evaluation moves past clean,
full-data test conditions: a single accuracy number is therefore a poor
proxy for trustworthy behavior. Calibration rankings are no exception, and
are in fact typically similar to, or less stable than, accuracy rankings:
clean-data calibration says little about calibration
under the shifts and data budgets a deployed model will actually meet. Broad benchmark evaluations therefore need to move beyond accuracy to offer a more holistic view of model performance.

\paragraph{Pretraining domain appears to matter for calibration under shift.}
On clean data the GeoFM-versus-general question shows no meaningful difference
(Section~\ref{sec:clean}), and under shift EO pretraining confers no accuracy
advantage either: pooled across the canon the EO encoders sit slightly below the
general-purpose ones, averaged over the three corruption families and
severities~1--5. The margin is small and varies by dataset, so we
do not read it as a general accuracy deficit, only as the absence of the
advantage EO pretraining is meant to provide.
Concurrent evidence from real-world EO shifts agrees.
EarthShift~\cite{doerksen2026earthshift}, which benchmarks eight GeoFMs across held-out geographies, sensors, and time periods, finds
that EO pretraining confers no robustness advantage over ImageNet-pretrained or
fully-supervised models, with all of them degrading by roughly 15 to 20 percent
out of distribution.
Calibration under shift is one attribute where pretraining \emph{domain} does
separate the selected models: the EO-pretrained encoders, under linear probing, carry the
larger ECE in all 15 severity-by-corruption cells
(Table~\ref{app:encoder_severity_full}, group rows), and the gap runs in the
overconfident direction throughout (Fig.~\ref{fig:signed_ece}). That makes the domain split
worth taking seriously not as a verdict on either group, but as an informative
axis for future work.

\paragraph{Competing explanations that require further study.}
The under-shift calibration gap is not cleanly attributable to pretraining
domain alone. Exclusion of cloudy imagery from EO pretraining corpora in many GeoFM training pipelines cannot explain the
gap, since it appears just as clearly under motion blur and Poisson--Gaussian
noise, corruptions unrelated to cloud. However, general augmentation exposure is harder to rule
out. Natural-image pretraining pipelines are augmentation-heavy in
exactly the relevant ways (Gaussian blur and photometric
jitter~\cite{oquab2023dinov2, simeoni2025dinov3}) while EO masked-reconstruction
pretraining relies mostly on geometric crop-and-flip and
masking~\cite{szwarcman2025prithvi, clay2024, Reed2022ScaleMAEAS}, so part of
the gap may reflect augmentation exposure rather than domain.
Within the EO group, reconstruction- and distillation-based encoders appear to
calibrate differently under shift, but with only two self-distillation encoders
among the 16 encoders this cannot be settled here and would need a purpose-built encoder
set. A further confound is structural rather than about pretraining data or
objective: the EO and general-purpose groups also differ in backbone
architecture, parameter count, and training recipe, so pretraining domain is not
varied in isolation. The identical probe isolates each encoder's representation,
but it cannot separate the effect of the pretraining corpus from that of
architecture and capacity; attributing the calibration gap to domain alone would
require a controlled study that holds architecture and capacity fixed while
varying only the pretraining data. Together these point to the natural follow-up: which pretraining objectives
yield robust, well-calibrated features, including joint-embedding predictive
(JEPA) objectives, which are absent from our encoder set; this is a question our analysis is
meant to open for further research directions.

\paragraph{Limitations of the frozen probing protocol.}
Every prediction we evaluate is produced by a trained linear head on top of the
frozen features, so the calibration behavior we report is a property of the
encoder and its head together, not of the encoder in isolation. The head is
nonetheless held fixed across all 16 encoders, with the same architecture,
protocol, and an accuracy-selected regularization strength never re-tuned on
calibration loss (Section~\ref{sec:probe}); it is therefore a shared constant,
and the between-encoder differences our claims rest on isolate the representation
rather than the head. What this coupling leaves open is whether the same encoders
would remain overconfident under a different head, in particular a fine-tuned
backbone: our evidence speaks to the frozen linear-probe regime that dominates
GeoFM deployment, and a richer head or full fine-tuning could in principle change
the calibration behavior we observe. Concurrent evidence suggests fine-tuning is
unlikely to remove the failure, since EarthShift~\cite{doerksen2026earthshift}
finds out-of-distribution accuracy largely invariant to fine-tuning strategy;
that result concerns accuracy rather than calibration and uses a different
evaluation protocol, so the fine-tuned regime remains open.

\paragraph{Mitigations have limits.}
Post-hoc temperature scaling, fit once on clean data as is standard practice,
cannot correct for shift it was never exposed to; ensembles help only partially at
$k\times$ cost, and a GP probe trades clean calibration for shift robustness
(Section~\ref{sec:mitigations}); confidence-based selective prediction cannot
defer around confident errors either (Section~\ref{sec:selective}). We do not
claim these are unfixable but rather point out that accessible tools often
used in the literature are not the easy fix to these issues and pre-training
strategies might have to be revisited.
A more promising direction may be to base correction and abstention on signals
other than the model's own confidence. Our selective-prediction result shows
that confidence fails precisely where it is needed
(Section~\ref{sec:selective}), whereas the representation itself carries
information about shift: the CKA and participation-ratio drift measures that
track the calibration failure in our mechanism analysis
(Section~\ref{sec:mechanism}) operate on the embedding rather than the softmax.
Exisiting literature of out-of-distribution detectors already focuses on activation based detectors rather
than from the softmax~\cite{yang2024generalized}, for instance by class-conditional
Mahalanobis distance in feature space~\cite{lee2018simple} or by distance to
nearest training neighbours~\cite{sun2022knnood}; TARDIS~\cite{ekim2025distribution}
is the EO-specific instance, built on internal activations and evaluated on EO
covariate and semantic shift. Whether such representation-level signals translate into
better-calibrated predictions under shift remains open future work.

\paragraph{Limitations.}
Our synthetic corruptions approximate but do not replace real EO shift;
curated real-shift benchmarks such as EarthShift~\cite{doerksen2026earthshift}
are complementary and important.
Our corruption suite spans three physically distinct mechanisms but is not
exhaustive: geometric transforms such as rotation, scale, and translation, along
with finer variation within each mechanism, are untested, and whether the
calibration failures extend to them is left to future work. The representation-drift analysis
(Section~\ref{sec:mechanism}) runs on the same four-dataset classification
canon as the core UQ results but at a single seed rather than five, and is
correlational, established across models rather than by intervention.
Segmentation is evaluated with a single frozen decoder architecture (a
lightweight FPN head~\cite{lin2017feature}; Section~\ref{sec:probe});
calibration behavior under shift may not transfer to decoders with richer
contextual modeling, e.g., UperNet's pyramid pooling~\cite{xiao2018unified}
or DeepLab-style spatial pyramid pooling~\cite{chen2018encoder}. More broadly, we never fine-tune the backbone, for classification
or segmentation (Section~\ref{sec:problem_setting}); whether the
calibration-under-shift failures we report persist after full fine-tuning
is untested and remains open, as discussed above. A severity grade is also not
one physical intensity applied identically across the canon. Cloud opacity is
tuned per dataset to the sensor's dynamic range, and the Poisson--Gaussian noise
parameters are set per sensor type (Section~\ref{sec:corruptions}), so
severity~3 denotes a comparable degree of degradation within a dataset rather
than an identical physical condition across datasets. Motion blur uses the same
kernel widths everywhere, but the datasets differ in ground sample distance, so
even there the same kernel corresponds to a different physical smear on the
ground. Finally, the downstream head is varied only
between a linear probe and a GP probe: we have no calibration-under-shift data
for richer heads such as an MLP, so the head-sensitivity conclusions are bounded
to these two.

\section{Conclusion}

EO benchmarks rank geospatial foundation models by accuracy on clean, full-data
test sets. Across 16 encoders, nine datasets, and two orthogonal stress axes, we show that this ranking does not hold once
evaluation moves beyond those conditions, and that calibration, a property
deployment usually requires, is no exception. Clean-data calibration says little
about calibration under shift or across data budgets, where we observe that the collapse is
consistently in the overconfident direction. We instead find a correlation of these metrics via a representational mechanism conditioned on pretraining domain. Generally our results demonstrate that GeoFM evaluation should span more
than one operating point: reporting a calibration or proper-scoring metric
alongside accuracy, and probing more than one stress axis (such as distribution
shift and training-data budget), makes visible how much a ranking depends on often implicit assumptions. The specific corruptions and budgets we use are
illustrative, not a prescription.

Concurrent work such as EarthShift~\cite{doerksen2026earthshift} and
REOBench~\cite{li2026reobench}, together with the results reported here, shows
that benchmark evaluation must move beyond a single accuracy measurement toward
a more holistic assessment. Such evaluation would give the community a richer
basis for judging which models are suitable for which tasks under which
conditions, and would help close the gap between benchmark results and
real-world deployment.

\section*{Acknowledgments}
The work of N. Lehmann and X. Zhu is supported by German Federal Ministry for Economic Affairs and Climate Action in the framework of the "national center of excellence ML4Earth" (grant number: 50EE2201C), by the Excellence Strategy of the Federal Government and the Länder through the TUM Innovation Network EarthCare, and by Munich Center for Machine Learning. The work of X. Zhu is also supported by the European Commission through the project “ThinkingEarth—Copernicus Foundation Models for a Thinking Earth” under the Horizon 2020 Research and Innovation program (Grant Agreement No. 101130544).

\appendices

\section{Corruption Examples}
\label{app:corruption_examples_sec}

\begin{figure*}[t]
  \centering
  \includegraphics[width=\textwidth]{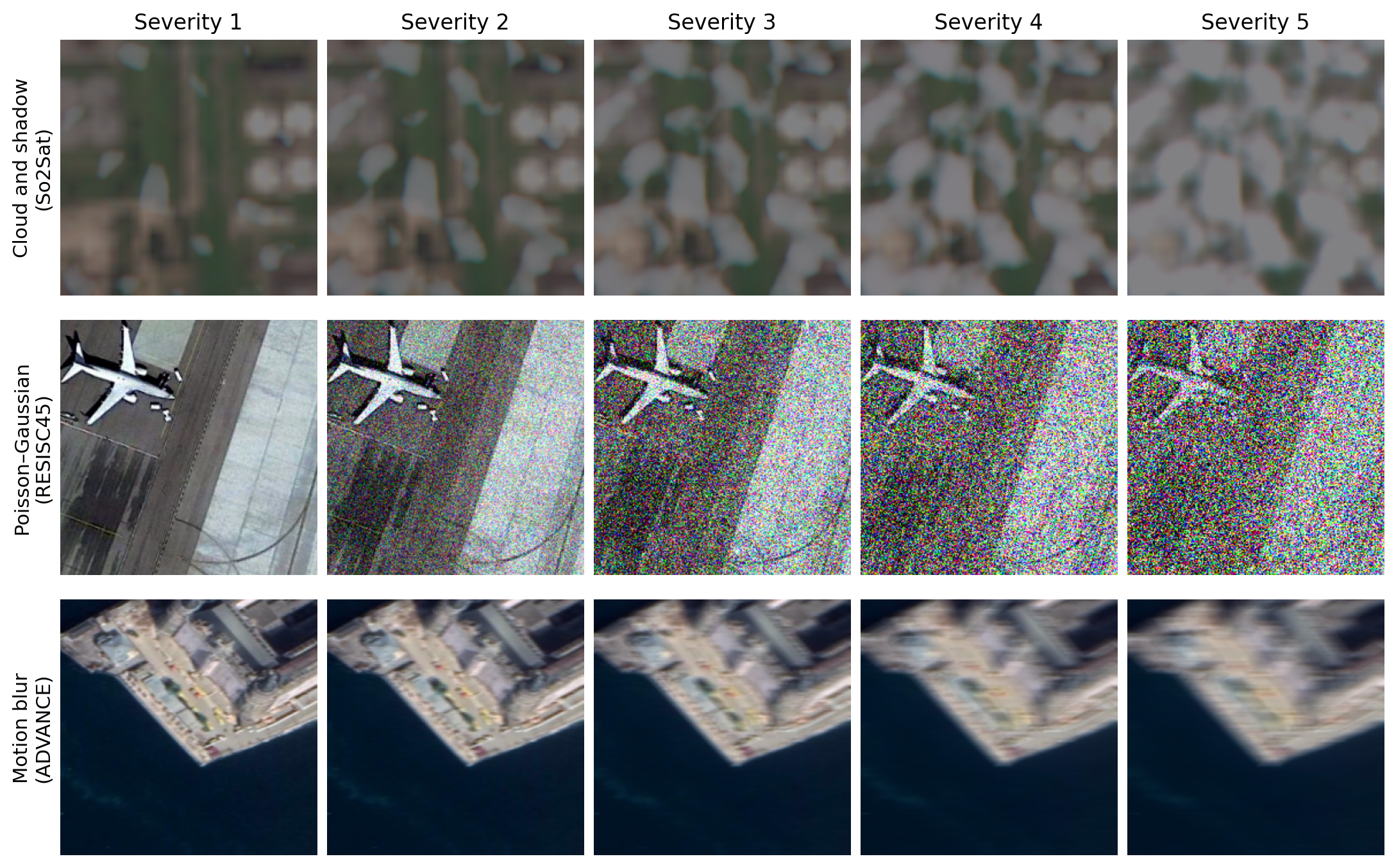}
  \caption{Synthetic corruptions at severities 1--5, one row per corruption
  family: cloud and shadow (top, So2Sat), Poisson--Gaussian sensor noise
  (middle, RESISC45), and motion blur (bottom, ADVANCE). Cloud coverage grows
  from light, localized patches to near-total occlusion; shot and readout noise
  intensify until fine structure is obscured; directional smearing increases
  with kernel width, degrading edges more gradually than cloud or sensor noise.}
  \label{app:corruption_examples}
\end{figure*}

Fig.~\ref{app:corruption_examples} gives a visual
sense of the three synthetic corruptions described in Section~\ref{sec:corruptions}
at each severity level, one representative image per corruption family shown on
three of the four classification datasets to illustrate the range of sensors and
scenes involved (So2Sat, RESISC45, and ADVANCE); the same corruption pipeline is
applied identically across all four datasets.

\section{Rank Sensitivity Under Shift (Scatter Views)}
\label{app:rank_sensitivity_sec}

\begin{figure*}[t]
  \centering
  \includegraphics[width=0.32\textwidth]{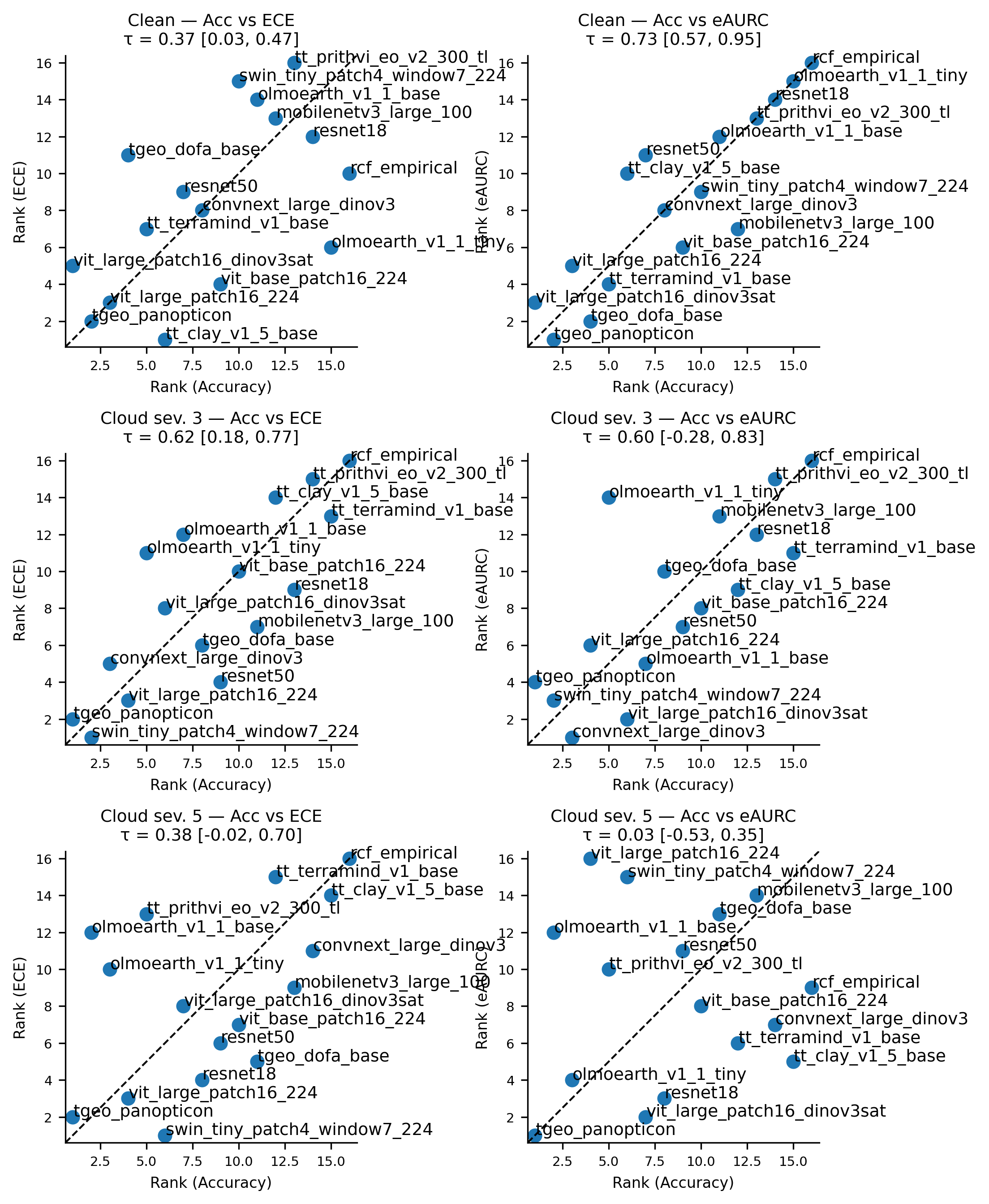}\hfill
  \includegraphics[width=0.32\textwidth]{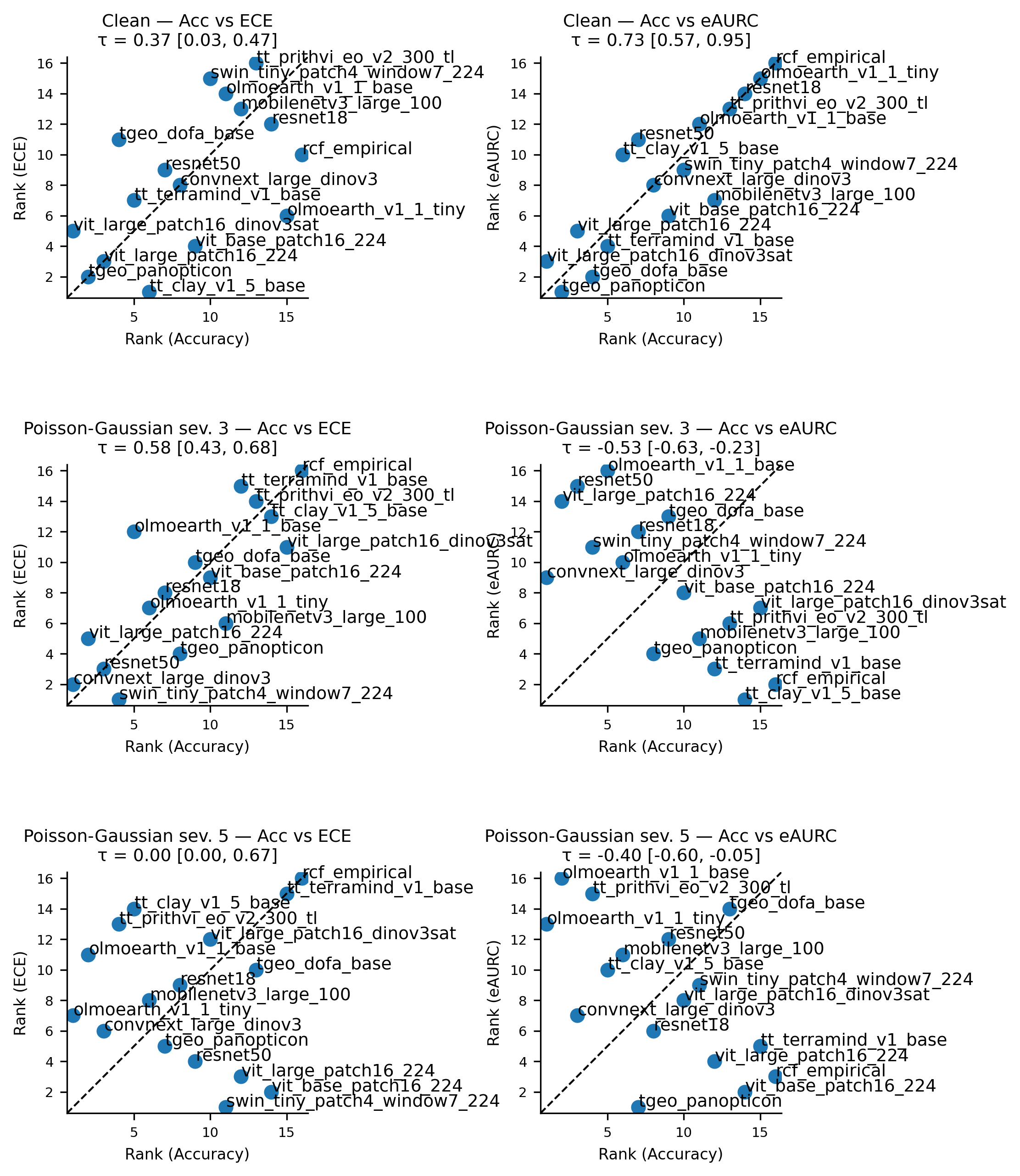}\hfill
  \includegraphics[width=0.32\textwidth]{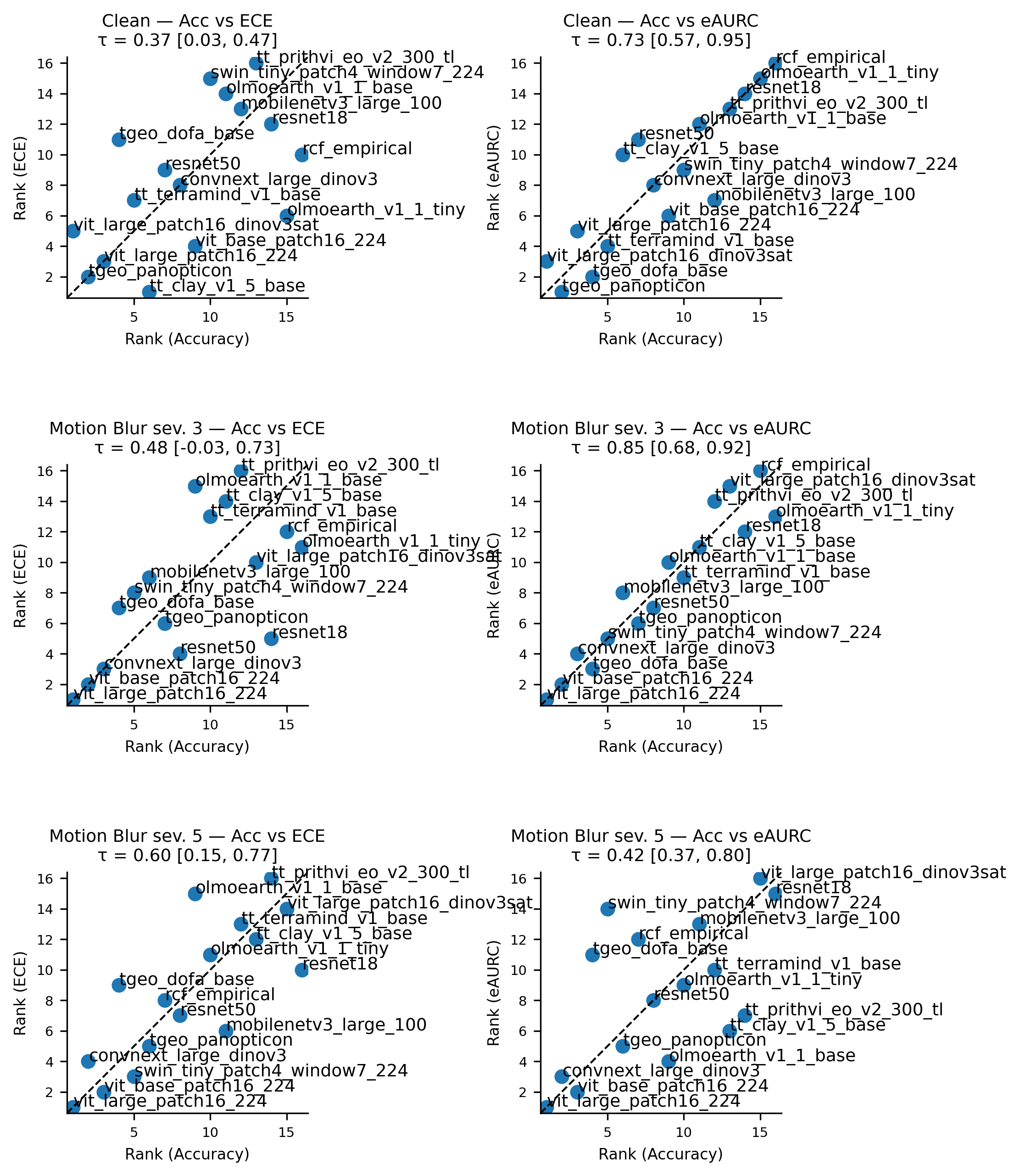}
  \caption{Per-model rank(accuracy) vs.\ rank(ECE) and rank(eAURC) at clean and
  two shifted severities, under cloud (left), Poisson--Gaussian (centre), and
  motion blur (right). Accuracy--ECE rank agreement is moderate across most
  conditions ($\tau\approx0.4$--$0.6$) but drops close to zero under severe
  Poisson--Gaussian noise ($\tau\approx0.0$ at severity~5); accuracy--eAURC rank
  agreement is high at clean ($\tau\approx0.73$) but collapses, and inverts,
  under cloud and Poisson--Gaussian corruption ($\tau$ as low as $-0.53$),
  while staying comparatively high under the milder motion-blur corruption.}
  \label{app:rank_sensitivity}
\end{figure*}

Fig.~\ref{app:rank_sensitivity} complements the rank-transfer curves in
Fig.~\ref{fig:tau_severity} with per-model scatters of rank(accuracy) against
rank(ECE) and rank(eAURC), at clean and two shifted severities, for each
corruption type. Points that leave the diagonal have accuracy and
calibration/selective-prediction rankings that disagree at that condition.

\section{Exclusion of SpaceNet7}
\label{app:seg}

Our segmentation canon uses five of the six datasets in the corruption sweep;
SpaceNet7 is excluded because on it the metric does not separate the encoders.
SpaceNet7 is two-class building-footprint segmentation on 4\,m PlanetScope
imagery, where buildings span only a few pixels at $224\times224$ and the
background class dominates by construction, so mIoU rewards a predictor that
attends to the background alone. Two properties of our results make that
concrete. First, capacity and pretraining buy nothing on this dataset: all 15
encoders fall within a $0.03$ mIoU band, and the $4$M-parameter MobileNetV3
($0.492$) matches or beats every GeoFM (e.g.\ OlmoEarth $0.479$, Prithvi
$0.470$). Fifteen encoders of widely differing capacity landing inside a $0.03$
band indicates that the metric is not separating them. Second, the
rank-stability metric rewards that degeneracy rather than penalizing it:
SpaceNet7 carries the \emph{highest} severity-1 Kendall $\tau$ of any
segmentation dataset under Poisson--Gaussian noise ($1.00$, against $0.98$ for
the next dataset) and the second highest under motion blur ($0.85$, behind
FLAIR2's $0.87$), and it stays positive at every severity of both, with a floor
of $0.16$, on a corruption where two of the five canon datasets have already
inverted at severity~1 (CloudSEN12 $-0.26$, FOTW $-0.28$). A ranking that barely
depends on the input is a reproducible ranking, so pooling SpaceNet7 would
inflate the aggregate rank stability this paper reports.

\section{Representation Drift: Supporting Analyses}
\label{app:drift}

\begin{figure*}[t]
  \centering
  \includegraphics[width=\textwidth]{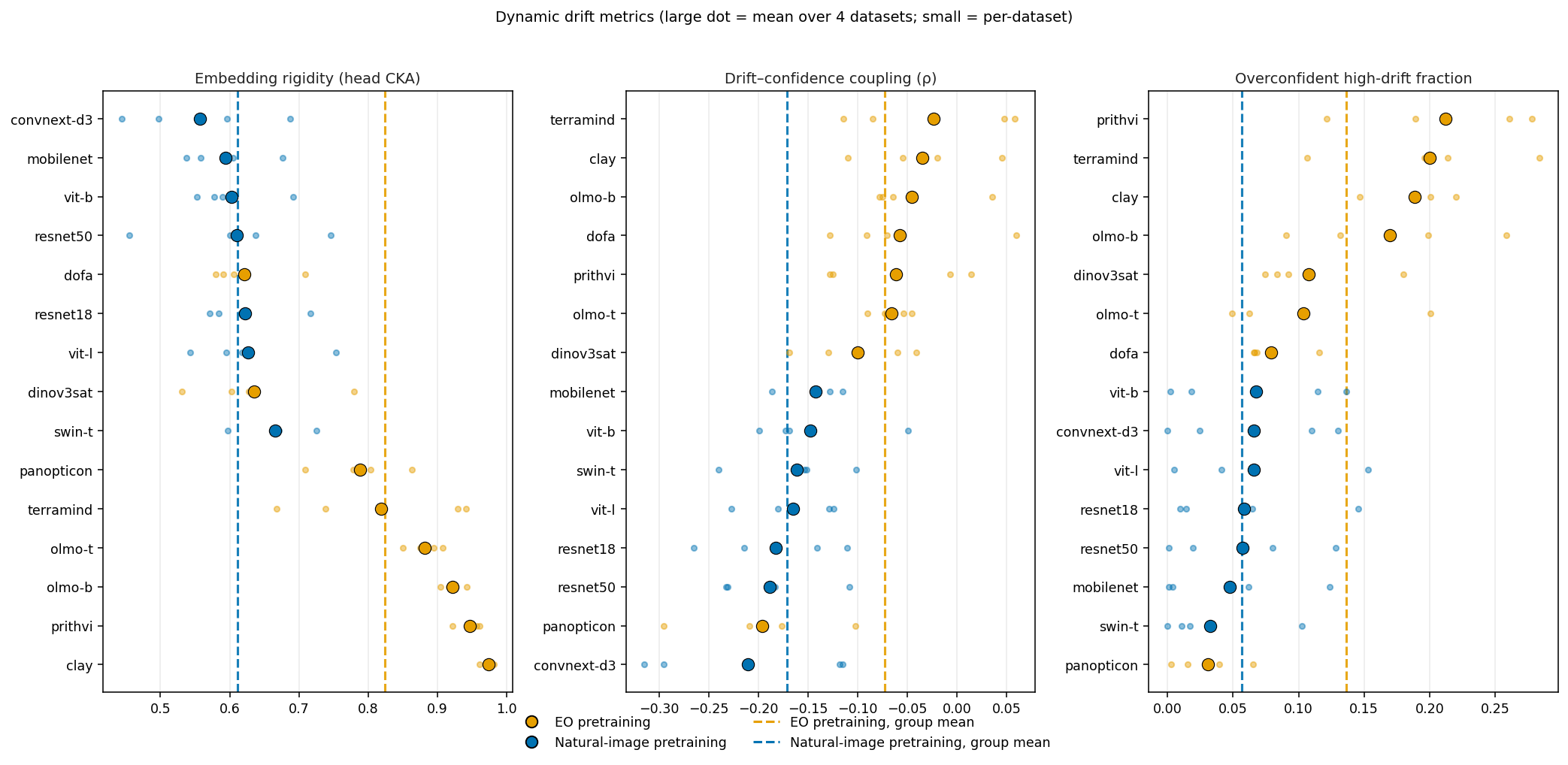}
  \caption{Per-model distributions of the three drift metrics (rigidity,
  drift--confidence coupling, overconfident high-drift fraction), sorted best
  to worst within each panel (classification, four-dataset canon); large dot =
  mean over datasets, small dots =
  per-dataset values (orange EO pretraining, blue natural-image pretraining).
  The two dashed vertical rules in each panel mark the group means, which are
  the values compared in Section~\ref{sec:mechanism}.
  EO-pretrained encoders occupy the worst end of all three panels on average,
  with Panopticon the clear exception: rigid yet the least overconfident encoder
  we evaluate.}
  \label{fig:track_b_dotplot}
\end{figure*}

\begin{figure}[t]
  \centering
  \includegraphics[width=\columnwidth]{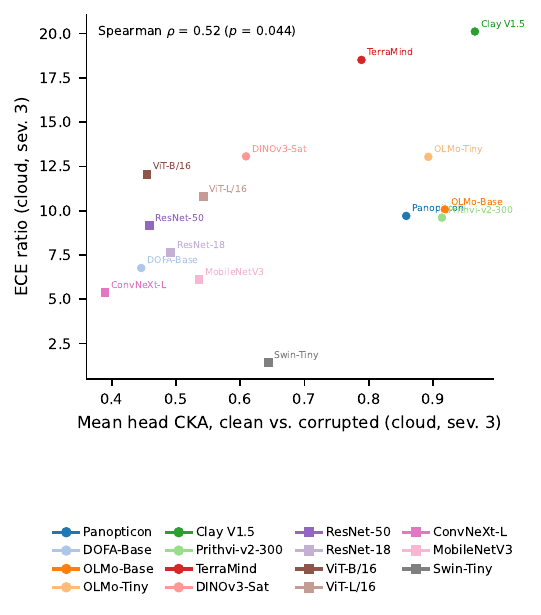}
  \caption{Head-layer CKA(clean, corrupted) versus ECE inflation at
  cloud severity~3, one point per encoder (classification, four-dataset
  canon; 15 encoders, RCF excluded for lack of a comparable layerwise CKA).
  Higher CKA (more rigid) is associated with higher ECE inflation
  (Spearman $\rho=0.52$, $p=.044$; a permutation test over $10^5$ draws gives
  $p=.048$): this illustrates the direct
  CKA--ECE association, not only through the composite overconfident-fraction
  metric in Fig.~\ref{fig:track_b_chain}. This is one of the 15
  severity-by-corruption conditions in Table~\ref{tab:cka_ece_robustness},
  which reports the same correlation for all of them.}
  \label{fig:cka_similarity_ece}
\end{figure}

\begin{figure}[t]
  \centering
  \includegraphics[width=\columnwidth]{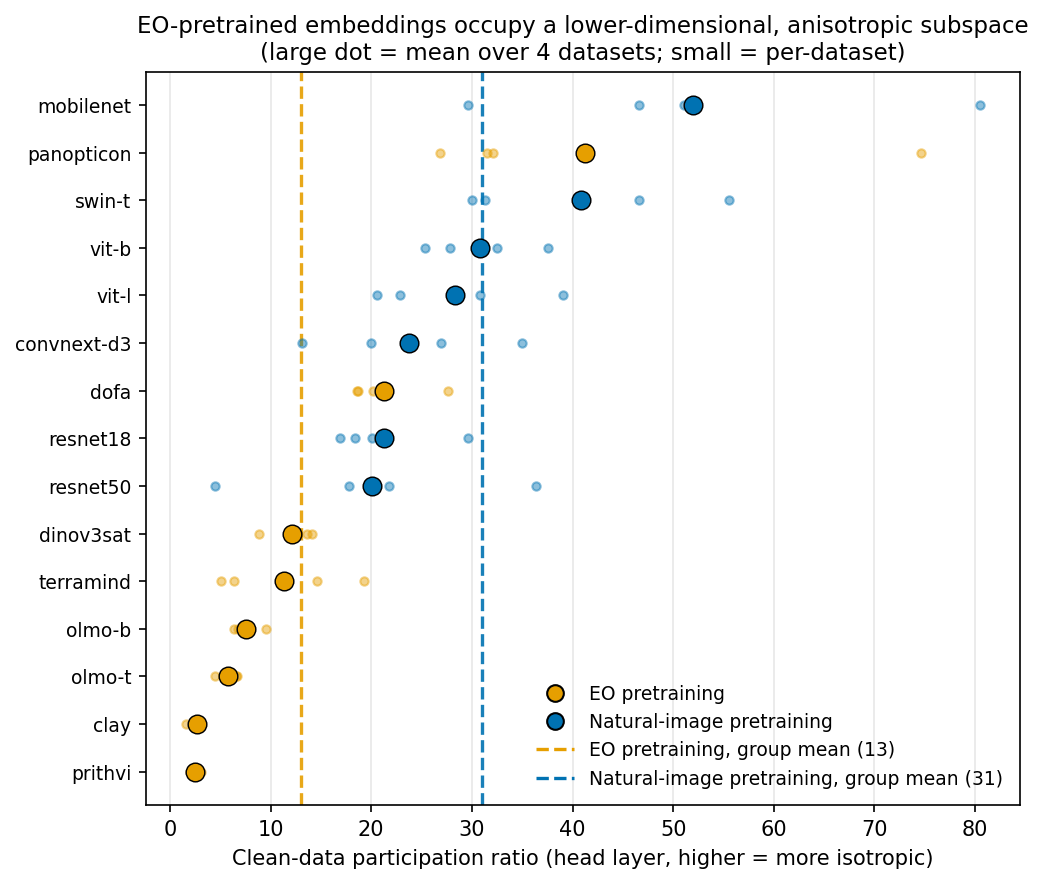}
  \caption{Clean-data participation ratio at the head layer, per model
  (large dot = mean over the four-dataset canon; small dots = per-dataset;
  dashed vertical rules = group means).
  On average, EO-pretrained encoders (orange) concentrate their variance in far
  fewer directions than natural-image encoders (blue)
  ($\approx13$ vs.\ $\approx31$), consistent with the (B1) rigidity claim, though
  Panopticon is an EO exception at the high-dimensional end.}
  \label{app:participation_ratio}
\end{figure}

Fig.~\ref{fig:track_b_dotplot} gives the per-model breakdown underlying the
(B1)--(B3) rigidity/coupling/overconfidence claims in
Section~\ref{sec:mechanism}. Fig.~\ref{fig:cka_similarity_ece} gives the
per-encoder scatter underlying the direct CKA$\to$ECE correlation reported
alongside Fig.~\ref{fig:cka_similarity_severity}.
Fig.~\ref{app:participation_ratio} shows the per-model
clean-data participation ratio underlying the (B1) rigidity claim.

\section{Per-Encoder Calibration Under Shift}
\label{app:encoder_severity_sec}

Table~\ref{app:encoder_severity_full} is the numerical backing for the
classification results in Sections~\ref{sec:clean} and~\ref{sec:shift}: clean
accuracy and clean ECE per encoder, then uncalibrated ECE at every severity of
every corruption family, averaged over the four-dataset canon. The four group
rows at the foot are means over the encoders of each pretraining domain; the two
RGB-only rows restrict the average to ADVANCE and RESISC45, the datasets on
which every encoder receives identical RGB input and no band configuration is
selected (Section~\ref{sec:encoders}). The EO-minus-general ECE gap is positive
in all 15 severity-by-corruption cells of the full-canon rows, and in 14 of the
15 RGB-only cells, the exception being Poisson--Gaussian severity~1, where the
two groups tie at $0.03$.

\begin{table*}[t]
  \centering
  \footnotesize
  \setlength{\tabcolsep}{2.5pt}
  \caption{Per-encoder clean accuracy and clean ECE, and uncalibrated ECE at
  severities 1--5 for each corruption family (classification, mean over the
  four-dataset canon). Encoders are ordered EO-pretrained, general-purpose, then
  the RCF baseline. The last four rows are group means over encoders: the two
  pretraining domains on the full canon, and the same two domains restricted to
  the RGB-only datasets (ADVANCE, RESISC45).}
  \label{app:encoder_severity_full}
  \begin{tabular}{lrrrrrrrrrrrrrrrrr}
\toprule
 & \multicolumn{2}{c}{Clean} & \multicolumn{5}{c}{Cloud} & \multicolumn{5}{c}{Poisson--Gauss.} & \multicolumn{5}{c}{Motion blur} \\
\cmidrule(lr){2-3}\cmidrule(lr){4-8}\cmidrule(lr){9-13}\cmidrule(lr){14-18}
Model & Acc. & ECE & 1 & 2 & 3 & 4 & 5 & 1 & 2 & 3 & 4 & 5 & 1 & 2 & 3 & 4 & 5 \\
\midrule
Panopticon & 0.881 & 0.023 & 0.04 & 0.15 & 0.23 & 0.28 & 0.35 & 0.19 & 0.29 & 0.44 & 0.52 & 0.57 & 0.03 & 0.03 & 0.08 & 0.15 & 0.20 \\
DOFA-Base & 0.867 & 0.039 & 0.08 & 0.17 & 0.26 & 0.32 & 0.38 & 0.23 & 0.38 & 0.60 & 0.68 & 0.65 & 0.04 & 0.04 & 0.08 & 0.17 & 0.25 \\
OLMo-Base & 0.837 & 0.044 & 0.14 & 0.33 & 0.44 & 0.52 & 0.62 & 0.23 & 0.60 & 0.71 & 0.75 & 0.79 & 0.05 & 0.09 & 0.21 & 0.34 & 0.41 \\
OLMo-Tiny & 0.775 & 0.026 & 0.10 & 0.25 & 0.34 & 0.44 & 0.52 & 0.25 & 0.55 & 0.60 & 0.60 & 0.58 & 0.03 & 0.05 & 0.15 & 0.23 & 0.30 \\
Clay V1.5 & 0.849 & 0.025 & 0.13 & 0.36 & 0.50 & 0.64 & 0.73 & 0.40 & 0.79 & 0.83 & 0.87 & 0.85 & 0.04 & 0.08 & 0.17 & 0.29 & 0.36 \\
Prithvi-v2-300 & 0.825 & 0.062 & 0.23 & 0.49 & 0.59 & 0.65 & 0.70 & 0.40 & 0.73 & 0.80 & 0.82 & 0.83 & 0.08 & 0.13 & 0.22 & 0.33 & 0.40 \\
TerraMind & 0.869 & 0.027 & 0.11 & 0.34 & 0.51 & 0.62 & 0.71 & 0.23 & 0.74 & 0.83 & 0.86 & 0.90 & 0.06 & 0.11 & 0.22 & 0.33 & 0.39 \\
DINOv3-Sat & 0.877 & 0.025 & 0.10 & 0.25 & 0.33 & 0.39 & 0.50 & 0.43 & 0.64 & 0.71 & 0.69 & 0.77 & 0.05 & 0.12 & 0.20 & 0.28 & 0.35 \\
\cmidrule(lr){1-18}
ResNet-50 & 0.829 & 0.030 & 0.14 & 0.22 & 0.27 & 0.32 & 0.39 & 0.22 & 0.33 & 0.44 & 0.53 & 0.57 & 0.03 & 0.04 & 0.08 & 0.14 & 0.20 \\
ResNet-18 & 0.810 & 0.039 & 0.15 & 0.24 & 0.30 & 0.34 & 0.33 & 0.26 & 0.39 & 0.56 & 0.67 & 0.65 & 0.04 & 0.04 & 0.08 & 0.19 & 0.28 \\
ViT-B/16 & 0.835 & 0.024 & 0.06 & 0.18 & 0.28 & 0.35 & 0.39 & 0.13 & 0.31 & 0.59 & 0.73 & 0.53 & 0.02 & 0.03 & 0.06 & 0.11 & 0.16 \\
ViT-L/16 & 0.870 & 0.020 & 0.04 & 0.13 & 0.21 & 0.31 & 0.35 & 0.10 & 0.24 & 0.43 & 0.60 & 0.58 & 0.02 & 0.03 & 0.04 & 0.08 & 0.12 \\
ConvNeXt-L & 0.854 & 0.048 & 0.07 & 0.14 & 0.26 & 0.36 & 0.52 & 0.07 & 0.16 & 0.28 & 0.39 & 0.54 & 0.05 & 0.06 & 0.09 & 0.14 & 0.18 \\
MobileNetV3 & 0.837 & 0.044 & 0.12 & 0.19 & 0.27 & 0.33 & 0.44 & 0.17 & 0.33 & 0.48 & 0.51 & 0.57 & 0.05 & 0.06 & 0.09 & 0.14 & 0.19 \\
Swin-Tiny & 0.835 & 0.064 & 0.06 & 0.08 & 0.09 & 0.19 & 0.32 & 0.07 & 0.20 & 0.25 & 0.41 & 0.43 & 0.07 & 0.07 & 0.09 & 0.15 & 0.18 \\
\cmidrule(lr){1-18}
RCF & 0.739 & 0.039 & 0.27 & 0.56 & 0.65 & 0.75 & 0.82 & 0.59 & 0.89 & 0.92 & 0.94 & 0.94 & 0.09 & 0.12 & 0.17 & 0.21 & 0.24 \\
\midrule
GeoFM mean & 0.847 & 0.034 & 0.12 & 0.29 & 0.40 & 0.49 & 0.56 & 0.29 & 0.59 & 0.69 & 0.73 & 0.74 & 0.05 & 0.08 & 0.17 & 0.26 & 0.33 \\
General mean & 0.839 & 0.038 & 0.09 & 0.17 & 0.24 & 0.32 & 0.39 & 0.14 & 0.28 & 0.43 & 0.55 & 0.55 & 0.04 & 0.05 & 0.07 & 0.14 & 0.19 \\
GeoFM mean (RGB only) & 0.884 & 0.023 & 0.12 & 0.35 & 0.44 & 0.52 & 0.60 & 0.03 & 0.51 & 0.70 & 0.76 & 0.77 & 0.05 & 0.12 & 0.27 & 0.43 & 0.52 \\
General mean (RGB only) & 0.917 & 0.024 & 0.07 & 0.18 & 0.25 & 0.33 & 0.39 & 0.03 & 0.17 & 0.40 & 0.50 & 0.55 & 0.03 & 0.04 & 0.09 & 0.20 & 0.29 \\
\bottomrule
\end{tabular}

\end{table*}

\section{Post-Hoc Calibration: Full Cloud Results}
\label{app:posthoc_full}

Table~\ref{app:posthoc_full_fig} reports $\Delta$ECE, $\Delta$NLL, and
$\Delta$Brier (method $-$ uncalibrated) for temperature scaling and deep
ensemble at cloud severities 1, 3, and 5, pooled across the four classification
datasets. Negative values indicate improvement; both methods leave large
residual miscalibration under shift, and Brier tracks the ECE trend, confirming
it is not a binning artifact. NLL does not: temperature scaling's $\Delta$NLL
stays negative through severity~5, where its $\Delta$ECE has already turned
positive by severity~4.

\begin{table*}[t]
  \centering
  \footnotesize
  \setlength{\tabcolsep}{4pt}
  \caption{Post-hoc calibration deltas under cloud corruption
  ($\Delta$ = method $-$ uncalibrated), pooled across the four classification
  datasets, at severities s1/s3/s5. Negative values indicate improvement; Brier
  tracks the ECE trend and NLL does not. RCF has no
  deep-ensemble variant.}
  \label{app:posthoc_full_fig}
  \begin{tabular}{llrrrrrrrrr}
\toprule
 &  & \multicolumn{3}{c}{$\Delta$ECE} & \multicolumn{3}{c}{$\Delta$NLL} & \multicolumn{3}{c}{$\Delta$Brier} \\
\cmidrule(lr){3-5}\cmidrule(lr){6-8}\cmidrule(lr){9-11}
Model & Method & s1 & s3 & s5 & s1 & s3 & s5 & s1 & s3 & s5 \\
\midrule
\multirow{2}{*}{Panopticon} & Temp. Scaling & +0.018 & +0.027 & +0.039 & +0.045 & +0.449 & +0.895 & +0.007 & +0.031 & +0.043 \\
 & Deep Ensemble & +0.000 & +0.005 & -0.015 & +0.013 & +0.115 & +0.149 & +0.003 & +0.005 & -0.011 \\
\multirow{2}{*}{DOFA-Base} & Temp. Scaling & -0.017 & +0.043 & +0.058 & -0.008 & +0.185 & +0.565 & -0.005 & +0.023 & +0.057 \\
 & Deep Ensemble & -0.002 & -0.015 & -0.007 & +0.004 & -0.030 & -0.078 & -0.002 & -0.017 & -0.014 \\
\multirow{2}{*}{OLMo-Base} & Temp. Scaling & -0.017 & -0.022 & -0.025 & -0.087 & -0.480 & -0.875 & -0.009 & -0.021 & -0.031 \\
 & Deep Ensemble & -0.072 & -0.093 & -0.182 & -0.284 & -2.685 & -4.672 & -0.017 & -0.111 & -0.222 \\
\multirow{2}{*}{OLMo-Tiny} & Temp. Scaling & +0.015 & +0.011 & +0.013 & +0.045 & +0.200 & +0.112 & +0.007 & +0.013 & +0.020 \\
 & Deep Ensemble & -0.081 & -0.077 & -0.122 & -0.213 & -0.731 & -1.969 & -0.024 & -0.050 & -0.082 \\
\multirow{2}{*}{Clay V1.5} & Temp. Scaling & +0.009 & +0.015 & +0.017 & +0.020 & +0.191 & +0.448 & +0.002 & +0.008 & +0.023 \\
 & Deep Ensemble & -0.015 & -0.051 & -0.153 & -0.044 & -0.585 & -1.476 & -0.000 & -0.039 & -0.185 \\
\multirow{2}{*}{Prithvi-v2-300} & Temp. Scaling & -0.065 & -0.066 & -0.057 & -0.904 & -3.282 & -4.936 & -0.042 & -0.089 & -0.080 \\
 & Deep Ensemble & -0.009 & -0.026 & +0.027 & -0.579 & -2.168 & -3.293 & +0.050 & -0.006 & +0.066 \\
\multirow{2}{*}{TerraMind} & Temp. Scaling & -0.001 & -0.006 & +0.000 & -0.005 & -0.109 & -0.546 & -0.000 & -0.007 & +0.000 \\
 & Deep Ensemble & -0.015 & -0.023 & -0.025 & -0.030 & -0.307 & -0.774 & -0.014 & -0.030 & -0.027 \\
\multirow{2}{*}{DINOv3-Sat} & Temp. Scaling & +0.014 & +0.016 & +0.017 & +0.052 & +0.068 & +0.049 & +0.013 & +0.020 & +0.022 \\
 & Deep Ensemble & -0.035 & -0.032 & -0.041 & -0.089 & -0.005 & -0.078 & -0.027 & -0.014 & -0.032 \\
\cmidrule(lr){1-11}
\multirow{2}{*}{ResNet-50} & Temp. Scaling & -0.012 & -0.025 & -0.026 & -0.041 & -0.170 & -0.305 & -0.003 & -0.017 & -0.030 \\
 & Deep Ensemble & -0.008 & -0.030 & -0.027 & -0.085 & -0.223 & -0.209 & -0.007 & -0.028 & -0.030 \\
\multirow{2}{*}{ResNet-18} & Temp. Scaling & +0.032 & +0.044 & +0.047 & +0.058 & +0.201 & +0.311 & +0.011 & +0.032 & +0.037 \\
 & Deep Ensemble & +0.002 & -0.018 & -0.013 & +0.026 & -0.067 & -0.034 & +0.004 & -0.017 & -0.011 \\
\multirow{2}{*}{ViT-B/16} & Temp. Scaling & -0.010 & -0.009 & -0.009 & -0.010 & -0.101 & -0.108 & -0.003 & -0.014 & -0.010 \\
 & Deep Ensemble & -0.002 & -0.010 & +0.008 & +0.016 & +0.004 & +0.053 & +0.005 & -0.014 & +0.008 \\
\multirow{2}{*}{ViT-L/16} & Temp. Scaling & +0.006 & +0.007 & +0.011 & -0.003 & -0.049 & -0.104 & -0.000 & -0.004 & +0.003 \\
 & Deep Ensemble & +0.016 & -0.012 & +0.007 & +0.004 & -0.072 & -0.010 & +0.003 & -0.012 & +0.016 \\
\multirow{2}{*}{ConvNeXt-L} & Temp. Scaling & -0.005 & +0.003 & +0.001 & -0.030 & -0.065 & -0.487 & -0.002 & +0.005 & -0.013 \\
 & Deep Ensemble & -0.002 & +0.007 & +0.015 & +0.066 & +0.170 & +0.801 & +0.012 & +0.006 & +0.008 \\
\multirow{2}{*}{MobileNetV3} & Temp. Scaling & -0.004 & -0.018 & -0.018 & -0.162 & -0.425 & -0.623 & -0.010 & -0.025 & -0.027 \\
 & Deep Ensemble & -0.014 & +0.006 & +0.020 & -0.136 & -0.251 & -0.051 & -0.018 & -0.006 & +0.023 \\
\multirow{2}{*}{Swin-Tiny} & Temp. Scaling & -0.011 & +0.069 & +0.083 & -0.022 & +0.112 & +0.517 & -0.006 & +0.023 & +0.077 \\
 & Deep Ensemble & +0.010 & +0.007 & +0.005 & -0.002 & +0.011 & -0.001 & -0.001 & +0.006 & +0.004 \\
\cmidrule(lr){1-11}
\multirow{1}{*}{RCF} & Temp. Scaling & +0.001 & +0.001 & +0.002 & +0.107 & +0.057 & +0.061 & +0.004 & +0.002 & +0.003 \\
\bottomrule
\end{tabular}

\end{table*}

\clearpage

\bibliographystyle{IEEEtran}
\bibliography{bibliography}

\begin{thebibliography}{10}
\providecommand{\url}[1]{#1}
\csname url@samestyle\endcsname
\providecommand{\newblock}{\relax}
\providecommand{\bibinfo}[2]{#2}
\providecommand{\BIBentrySTDinterwordspacing}{\spaceskip=0pt\relax}
\providecommand{\BIBentryALTinterwordstretchfactor}{4}
\providecommand{\BIBentryALTinterwordspacing}{\spaceskip=\fontdimen2\font plus
\BIBentryALTinterwordstretchfactor\fontdimen3\font minus
  \fontdimen4\font\relax}
\providecommand{\BIBforeignlanguage}[2]{{%
\expandafter\ifx\csname l@#1\endcsname\relax
\typeout{** WARNING: IEEEtran.bst: No hyphenation pattern has been}%
\typeout{** loaded for the language `#1'. Using the pattern for}%
\typeout{** the default language instead.}%
\else
\language=\csname l@#1\endcsname
\fi
#2}}
\providecommand{\BIBdecl}{\relax}
\BIBdecl

\bibitem{bommasani2021opportunities}
R.~Bommasani, D.~A. Hudson, E.~Adeli, R.~Altman, S.~Arora, S.~von Arx, M.~S.
  Bernstein, J.~Bohg, A.~Bosselut, E.~Brunskill \emph{et~al.}, ``On the
  opportunities and risks of foundation models,'' \emph{arXiv preprint
  arXiv:2108.07258}, 2021.

\bibitem{brown2020language}
T.~Brown, B.~Mann, N.~Ryder, M.~Subbiah, J.~D. Kaplan, P.~Dhariwal,
  A.~Neelakantan, P.~Shyam, G.~Sastry, A.~Askell \emph{et~al.}, ``Language
  models are few-shot learners,'' \emph{Advances in Neural Information
  Processing Systems (NeurIPS)}, 2020.

\bibitem{rombach2022high}
R.~Rombach, A.~Blattmann, D.~Lorenz, P.~Esser, and B.~Ommer, ``High-resolution
  image synthesis with latent diffusion models,'' in \emph{IEEE/CVF Conference
  on Computer Vision and Pattern Recognition (CVPR)}, 2022.

\bibitem{ho2022video}
J.~Ho, T.~Salimans, A.~Gritsenko, W.~Chan, M.~Norouzi, and D.~J. Fleet, ``Video
  diffusion models,'' \emph{Advances in Neural Information Processing Systems
  (NeurIPS)}, 2022.

\bibitem{huang2025survey}
Z.~Huang, H.~Yan, Q.~Zhan, S.~Yang, M.~Zhang, C.~Zhang, Y.~Lei, Z.~Liu, Q.~Liu,
  and Y.~Wang, ``A survey on remote sensing foundation models: From vision to
  multimodality,'' \emph{arXiv preprint arXiv:2503.22081}, 2025.

\bibitem{zhu2026foundations}
X.~X. Zhu, Z.~Xiong, Y.~Wang, A.~J. Stewart, K.~Heidler, Y.~Wang, Z.~Yuan,
  T.~Dujardin, Q.~Xu, and Y.~Shi, ``On the foundations of earth foundation
  models,'' \emph{Communications Earth \& Environment}, 2026.

\bibitem{corley2026no}
I.~Corley, N.~Lehmann, C.~Robinson, G.~Tseng, A.~Fuller, H.~Alemohammad,
  E.~Shelhamer, J.~Marcus, and H.~Kerner, ``No one knows the state of the art
  in geospatial foundation models,'' \emph{arXiv preprint arXiv:2605.12678},
  2026.

\bibitem{mehrtash2020confidence}
A.~Mehrtash, W.~M. Wells, C.~M. Tempany, P.~Abolmaesumi, and T.~Kapur,
  ``Confidence calibration and predictive uncertainty estimation for deep
  medical image segmentation,'' \emph{IEEE Transactions on Medical Imaging},
  2020.

\bibitem{ovadia2019can}
Y.~Ovadia, E.~Fertig, J.~Ren, Z.~Nado, D.~Sculley, S.~Nowozin, J.~Dillon,
  B.~Lakshminarayanan, and J.~Snoek, ``Can you trust your model's uncertainty?
  evaluating predictive uncertainty under dataset shift,'' \emph{Advances in
  neural information processing systems}, vol.~32, 2019.

\bibitem{hendrycks2021many}
D.~Hendrycks, S.~Basart, N.~Mu, S.~Kadavath, F.~Wang, E.~Dorundo, R.~Desai,
  T.~Zhu, S.~Parajuli, M.~Guo \emph{et~al.}, ``The many faces of robustness: A
  critical analysis of out-of-distribution generalization,'' in
  \emph{Proceedings of the IEEE/CVF international conference on computer
  vision}, 2021, pp. 8340--8349.

\bibitem{moreno2012unifying}
J.~G. Moreno-Torres, T.~Raeder, R.~Alaiz-Rodr{\'\i}guez, N.~V. Chawla, and
  F.~Herrera, ``A unifying view on dataset shift in classification,''
  \emph{Pattern Recognition}, vol.~45, no.~1, pp. 521--530, 2012.

\bibitem{quinonero2009dataset}
J.~Qui{\~n}onero-Candela, M.~Sugiyama, A.~Schwaighofer, and N.~D. Lawrence,
  \emph{Dataset Shift in Machine Learning}.\hskip 1em plus 0.5em minus
  0.4em\relax Cambridge, MA: MIT Press, 2008.

\bibitem{tamang2025handling}
L.~Tamang, M.~R. Bouadjenek, R.~Dazeley, and S.~Aryal, ``Handling
  out-of-distribution data: A survey,'' \emph{IEEE Transactions on Knowledge
  and Data Engineering}, 2025, arXiv:2507.21160.

\bibitem{doerksen2026earthshift}
K.~Doerksen and H.~Kerner, ``Earthshift: a benchmark for measuring robustness
  to real-world distribution shifts in earth observation,'' \emph{arXiv
  preprint arXiv:2605.29330}, 2026.

\bibitem{guo2017calibration}
C.~Guo, G.~Pleiss, Y.~Sun, and K.~Q. Weinberger, ``On calibration of modern
  neural networks,'' in \emph{International Conference on Machine Learning},
  2017, pp. 1321--1330.

\bibitem{mucsanyi2023trustworthy}
B.~Mucs{\'a}nyi, M.~Kirchhof, E.~Nguyen, A.~Rubinstein, and S.~J. Oh,
  ``Trustworthy machine learning,'' \emph{arXiv preprint arXiv:2310.08215},
  2023.

\bibitem{liu2021trustworthy}
H.~Liu, Y.~Wang, W.~Fan, X.~Liu, Y.~Li, S.~Jain, Y.~Liu, A.~K. Jain, and
  J.~Tang, ``Trustworthy {AI}: A computational perspective,'' \emph{ACM
  Transactions on Intelligent Systems and Technology}, 2022.

\bibitem{gawlikowski2023survey}
J.~Gawlikowski, C.~R.~N. Tassi, M.~Ali, J.~Lee, M.~Humt, J.~Feng, A.~Kruspe,
  R.~Triebel, P.~Jung, R.~Roscher \emph{et~al.}, ``A survey of uncertainty in
  deep neural networks,'' \emph{Artificial intelligence review}, vol.~56, no.
  Suppl 1, pp. 1513--1589, 2023.

\bibitem{dominique2025role}
B.~Dominique, P.~Lam, N.~Kurtansky, J.~Weber, K.~Kose, V.~Rotemberg, and J.~Dy,
  ``On the role of calibration in benchmarking algorithmic fairness for skin
  cancer detection,'' \emph{arXiv preprint arXiv:2511.07700}, 2025.

\bibitem{ju2024monica}
L.~Ju, S.~Yan, Y.~Zhou, Y.~Nan, X.~Xing, P.~Duan, and Z.~Ge, ``Monica:
  Benchmarking on long-tailed medical image classification,'' \emph{arXiv
  preprint arXiv:2410.02010}, 2024.

\bibitem{kadavath2022language}
S.~Kadavath, T.~Conerly, A.~Askell, T.~Henighan, D.~Drain, E.~Perez,
  N.~Schiefer, Z.~{Hatfield-Dodds}, N.~{DasSarma}, E.~{Tran-Johnson},
  S.~Johnston, S.~{El-Showk}, A.~Jones, N.~Elhage, T.~Hume, A.~Chen, Y.~Bai,
  S.~Bowman, S.~Fort, D.~Ganguli, D.~Hernandez, J.~Jacobson, J.~Kernion,
  S.~Kravec, L.~Lovitt, K.~Ndousse, C.~Olsson, S.~Ringer, D.~Amodei, T.~Brown,
  J.~Clark, N.~Joseph, B.~Mann, S.~{McCandlish}, C.~Olah, and J.~Kaplan,
  ``Language models (mostly) know what they know,'' \emph{arXiv preprint
  arXiv:2207.05221}, 2022.

\bibitem{center2026benchmark}
L.~Phan, A.~Gatti, Z.~Han, N.~Li, D.~Hendrycks \emph{et~al.}, ``A benchmark of
  expert-level academic questions to assess {AI} capabilities,'' \emph{Nature},
  vol. 649, no. 8099, pp. 1139--1146, 2026.

\bibitem{minderer2021revisiting}
M.~Minderer, J.~Djolonga, R.~Romijnders, F.~Hubis, X.~Zhai, N.~Houlsby,
  D.~Tran, and M.~Lucic, ``Revisiting the calibration of modern neural
  networks,'' \emph{Advances in neural information processing systems},
  vol.~34, pp. 15\,682--15\,694, 2021.

\bibitem{tao2024benchmark}
L.~Tao, Y.~Zhu, H.~Guo, M.~Dong, and C.~Xu, ``A benchmark study on
  calibration,'' in \emph{International Conference on Learning
  Representations}, vol. 2024, 2024, pp. 25\,084--25\,096.

\bibitem{lacoste2023geo}
A.~Lacoste, N.~Lehmann, P.~Rodriguez, E.~Sherwin, H.~Kerner, B.~L{\"u}tjens,
  J.~Irvin, D.~Dao, H.~Alemohammad, A.~Drouin \emph{et~al.}, ``Geo-bench:
  Toward foundation models for earth monitoring,'' \emph{Advances in Neural
  Information Processing Systems}, vol.~36, pp. 51\,080--51\,093, 2023.

\bibitem{simumba2025geo}
\BIBentryALTinterwordspacing
N.~Simumba, N.~Lehmann, P.~Fraccaro, H.~Alemohammad, G.~De~Mel, S.~Khan,
  M.~Maskey, N.~Long{\'e}p{\'e}, X.~X. Zhu, H.~Kerner, J.~Bernabe~Moreno, and
  A.~Lacoste, ``{GEO}-bench-2: From performance to capability, rethinking
  evaluation in geospatial {AI},'' \emph{Transactions on Machine Learning
  Research}, 2026. [Online]. Available:
  \url{https://openreview.net/forum?id=NPf175jnP1}
\BIBentrySTDinterwordspacing

\bibitem{Marsocci2024PANGAEAAG}
V.~Marsocci, Y.~Jia, G.~L. Bellier, D.~Kerekes, L.~Zeng, S.~Hafner, S.~Gerard,
  E.~Brune, R.~Yadav, A.~Shibli, H.~Fang, Y.~Ban, M.~Vergauwen, N.~Audebert,
  and A.~Nascetti, ``Pangaea: A global and inclusive benchmark for geospatial
  foundation models,'' \emph{ArXiv}, vol. abs/2412.04204, 2024.

\bibitem{li2026reobench}
X.~Li, Y.~Tao, S.~Zhang, S.~Liu, Z.~Xiong, C.~Luo, L.~Liu, M.~Pechenizkiy,
  X.~Zhu, and T.~Huang, ``Reobench: Benchmarking robustness of earth
  observation foundation models,'' \emph{Advances in Neural Information
  Processing Systems}, vol.~38, 2025, datasets and Benchmarks Track.

\bibitem{Klemmer2023SatCLIPGG}
K.~Klemmer, E.~Rolf, C.~Robinson, L.~Mackey, and M.~Rußwurm, ``Satclip:
  Global, general-purpose location embeddings with satellite imagery,''
  \emph{ArXiv}, vol. abs/2311.17179, 2023.

\bibitem{Prexl2024SenPaMAESP}
J.~Prexl and M.~Schmitt, ``Senpa-mae: Sensor parameter aware masked autoencoder
  for multi-satellite self-supervised pretraining,'' \emph{ArXiv}, vol.
  abs/2408.11000, 2024.

\bibitem{Braham2024SpectralEarthTH}
N.~Braham, C.~Albrecht, J.~Mairal, J.~Chanussot, Y.~Wang, and X.~X. Zhu,
  ``Spectralearth: Training hyperspectral foundation models at scale,''
  \emph{IEEE Journal of Selected Topics in Applied Earth Observations and
  Remote Sensing}, vol.~18, pp. 16\,780--16\,797, 2025.

\bibitem{waldmann2025panopticon}
L.~Waldmann, A.~Shah, Y.~Wang, N.~Lehmann, A.~Stewart, Z.~Xiong, X.~X. Zhu,
  S.~Bauer, and J.~Chuang, ``Panopticon: Advancing any-sensor foundation models
  for earth observation,'' in \emph{Proceedings of the Computer Vision and
  Pattern Recognition Conference}, 2025, pp. 2204--2214.

\bibitem{Gong2024CrossEarthGV}
Z.~Gong, Z.~Wei, D.~Wang, X.~Hu, X.~Ma, H.~Chen, Y.~Jia, Y.~Deng, Z.~Ji,
  X.~Zhu, X.~Yang, N.~Yokoya, J.~Zhang, B.~Du, J.~Yan, and L.~Zhang,
  ``Crossearth: Geospatial vision foundation model for domain generalizable
  remote sensing semantic segmentation,'' \emph{IEEE Transactions on Pattern
  Analysis and Machine Intelligence}, vol.~48, pp. 5147--5164, 2026.

\bibitem{Reed2022ScaleMAEAS}
C.~Reed, R.~Gupta, S.~Li, S.~Brockman, C.~Funk, B.~Clipp, K.~Keutzer,
  S.~Candido, M.~Uyttendaele, and T.~Darrell, ``Scale-mae: A scale-aware masked
  autoencoder for multiscale geospatial representation learning,'' \emph{2023
  IEEE/CVF International Conference on Computer Vision (ICCV)}, pp. 4065--4076,
  2023.

\bibitem{Tang2024CrossScaleMA}
M.~Tang, A.~Cozma, K.~Georgiou, and H.~Qi, ``Cross-scale mae: A tale of
  multi-scale exploitation in remote sensing,'' \emph{ArXiv}, vol.
  abs/2401.15855, 2024.

\bibitem{Feng2025TESSERATE}
Z.~Feng, C.~Atzberger, S.~Jaffer, J.~Knezevic, S.~Sormunen, R.~Young, M.~C.
  Lisaius, M.~Immitzer, T.~Jackson, J.~Ball, D.~A. Coomes, A.~Madhavapeddy,
  A.~Blake, and S.~Keshav, ``Tessera: Temporal embeddings of surface spectra
  for earth representation and analysis,'' in \emph{Proceedings of the IEEE/CVF
  Conference on Computer Vision and Pattern Recognition (CVPR)}, Jun. 2026, pp.
  34\,818--34\,831.

\bibitem{Wang2024HyperSIGMAHI}
D.~Wang, M.~Hu, Y.~Jin, Y.~Miao, J.~Yang, Y.~Xu, X.~Qin, J.~Ma, L.~Sun, C.~Li,
  C.~Fu, H.~Chen, C.~Han, N.~Yokoya, J.~Zhang, M.~Xu, L.~Liu, L.~Zhang, C.~wu,
  B.~Du, D.~Tao, and L.~Zhang, ``Hypersigma: Hyperspectral intelligence
  comprehension foundation model,'' \emph{IEEE Transactions on Pattern Analysis
  and Machine Intelligence}, vol.~47, pp. 6427--6444, 2025.

\bibitem{ekim2025distribution}
B.~Ekim, G.~A. Tadesse, C.~Robinson, G.~Hacheme, M.~Schmitt, R.~Dodhia, and
  J.~M. Lavista~Ferres, ``Distribution shifts at scale: Out-of-distribution
  detection in earth observation,'' in \emph{Proceedings of the IEEE/CVF
  Conference on Computer Vision and Pattern Recognition Workshops}, 2025.

\bibitem{Chen2021SelfsupervisedRS}
Y.~Chen and L.~Bruzzone, ``Self-supervised remote sensing images change
  detection at pixel-level,'' 2021.

\bibitem{kendall2017uncertainties}
A.~Kendall and Y.~Gal, ``What uncertainties do we need in bayesian deep
  learning for computer vision?'' \emph{Advances in neural information
  processing systems}, vol.~30, 2017.

\bibitem{gonzalezcalabuig2026shrugfm}
M.~Gonzalez-Calabuig, K.-H. Cohrs, V.~Nedungadi, Z.~Osika, R.~Cartuyvels,
  S.~Knoblauch, J.~Massant, S.~Nath, P.~Ebel, and V.~Sitokonstantinou,
  ``Shrug-fm: Reliability-aware foundation models for earth observation,'' in
  \emph{Proceedings of the IEEE/CVF Conference on Computer Vision and Pattern
  Recognition (CVPR) Workshops (EarthVision)}, Jun. 2026, pp. 7919--7928.

\bibitem{alphaearth2025}
C.~F. Brown, M.~R. Kazmierski, V.~J. Pasquarella, W.~J. Rucklidge,
  M.~Samsikova, C.~Zhang, E.~Shelhamer, E.~Lahera, O.~Wiles, S.~Ilyushchenko,
  N.~Gorelick, L.~L. Zhang, S.~Alj, E.~Schechter, S.~Askay, O.~Guinan,
  R.~Moore, A.~Boukouvalas, and P.~Kohli, ``{AlphaEarth} foundations: An
  embedding field model for accurate and efficient global mapping from sparse
  label data,'' \emph{arXiv preprint arXiv:2507.22291}, 2025.

\bibitem{Danish2025TerraFMAS}
M.~S. Danish, M.~A. Munir, S.~R.~A. Shah, M.~H. Khan, R.~Anwer, J.~Laaksonen,
  F.~Khan, and S.~H. Khan, ``Terrafm: A scalable foundation model for unified
  multisensor earth observation,'' \emph{ArXiv}, vol. abs/2506.06281, 2025.

\bibitem{lakshminarayanan2017simple}
B.~Lakshminarayanan, A.~Pritzel, and C.~Blundell, ``Simple and scalable
  predictive uncertainty estimation using deep ensembles,'' in \emph{Advances
  in Neural Information Processing Systems}, vol.~30, 2017.

\bibitem{Angelopoulos2021AGI}
A.~N. Angelopoulos and S.~Bates, ``A gentle introduction to conformal
  prediction and distribution-free uncertainty quantification,'' \emph{ArXiv},
  vol. abs/2107.07511, 2021.

\bibitem{lin2017feature}
T.-Y. Lin, P.~Doll{\'a}r, R.~Girshick, K.~He, B.~Hariharan, and S.~Belongie,
  ``Feature pyramid networks for object detection,'' in \emph{Proceedings of
  the IEEE Conference on Computer Vision and Pattern Recognition}, 2017, pp.
  2117--2125.

\bibitem{aybar2022cloudsen12}
C.~Aybar, L.~Ysuhuaylas, J.~Loja \emph{et~al.}, ``Cloudsen12, a global dataset
  for semantic understanding of cloud and cloud shadow in sentinel-2,''
  \emph{Scientific Data}, vol.~9, no.~1, p. 782, 2022.

\bibitem{toker2022dynamicearthnet}
A.~Toker, L.~Kondmann, M.~Weber \emph{et~al.}, ``Dynamicearthnet: Daily
  multi-spectral satellite dataset for semantic change segmentation,'' in
  \emph{Proceedings of the IEEE/CVF Conference on Computer Vision and Pattern
  Recognition (CVPR)}, 2022.

\bibitem{garioud2023flair}
A.~Garioud, N.~Gonthier, L.~Landrieu \emph{et~al.}, ``Flair: a country-scale
  land cover semantic segmentation dataset from multi-source optical imagery,''
  in \emph{Advances in Neural Information Processing Systems (NeurIPS) Datasets
  and Benchmarks Track}, 2023.

\bibitem{kerner2025fields}
\BIBentryALTinterwordspacing
H.~Kerner, S.~Chaudhari, A.~Ghosh, C.~Robinson, A.~Ahmad, E.~Choi, N.~Jacobs,
  C.~Holmes, M.~Mohr, R.~Dodhia, J.~M. Lavista~Ferres, and J.~Marcus, ``Fields
  of the world: A machine learning benchmark dataset for global agricultural
  field boundary segmentation,'' \emph{Proceedings of the AAAI Conference on
  Artificial Intelligence}, vol.~39, no.~27, pp. 28\,151--28\,159, Apr. 2025.
  [Online]. Available:
  \url{https://ojs.aaai.org/index.php/AAAI/article/view/35034}
\BIBentrySTDinterwordspacing

\bibitem{vanetten2018spacenet}
A.~Van~Etten, D.~Lindenbaum, and T.~M. Bacastow, ``Spacenet: A remote sensing
  dataset and challenge series,'' \emph{arXiv preprint arXiv:1807.01232}, 2018.

\bibitem{hu2020cross}
D.~Hu, X.~Li, L.~Mou, P.~Jin, D.~Chen, L.~Jing, X.~Zhu, and D.~Dou,
  ``Cross-task transfer for geotagged audiovisual aerial scene recognition,''
  in \emph{European conference on computer vision}.\hskip 1em plus 0.5em minus
  0.4em\relax Springer, 2020, pp. 68--84.

\bibitem{cheng2017remote}
G.~Cheng, J.~Han, and X.~Lu, ``Remote sensing image scene classification:
  Benchmark and state of the art,'' \emph{Proceedings of the IEEE}, vol. 105,
  no.~10, pp. 1865--1883, 2017.

\bibitem{zhu2020so2sat}
X.~X. Zhu, J.~Hu, C.~Qiu, Y.~Shi, J.~Kang, L.~Mou, H.~Bagheri, M.~Haberle,
  Y.~Hua, R.~Huang \emph{et~al.}, ``So2sat lcz42: A benchmark data set for the
  classification of global local climate zones [software and data sets],''
  \emph{IEEE Geoscience and Remote Sensing Magazine}, vol.~8, no.~3, pp.
  76--89, 2020.

\bibitem{rolf2021generalizable}
E.~Rolf, J.~Proctor, T.~Carleton, I.~Bolliger, V.~Shankar, M.~Ishihara,
  B.~Recht, and S.~Hsiang, ``A generalizable and accessible approach to machine
  learning with global satellite imagery,'' \emph{Nature communications},
  vol.~12, no.~1, p. 4392, 2021.

\bibitem{corley2024revisiting}
I.~Corley, C.~Robinson, R.~Dodhia, J.~M.~L. Ferres, and P.~Najafirad,
  ``Revisiting pre-trained remote sensing model benchmarks: resizing and
  normalization matters,'' in \emph{Proceedings of the IEEE/CVF Conference on
  Computer Vision and Pattern Recognition}, 2024, pp. 3162--3172.

\bibitem{clay2024}
{Clay Foundation}, ``Clay foundation model,''
  \url{https://github.com/Clay-foundation/model}, 2024.

\bibitem{szwarcman2025prithvi}
D.~Szwarcman, S.~Roy, P.~Fraccaro, O.~E. G{\'\i}slason, B.~Blumenstiel,
  R.~Ghosal, P.~H. De~Oliveira, J.~L. de~Sousa~Almeida, R.~Sedona, Y.~Kang
  \emph{et~al.}, ``Prithvi-eo-2.0: A versatile multi-temporal foundation model
  for earth observation applications,'' \emph{IEEE Transactions on Geoscience
  and Remote Sensing}, 2025.

\bibitem{xiong2024neural}
Z.~Xiong, Y.~Wang, F.~Zhang, A.~J. Stewart, J.~Hanna, D.~Borth, I.~Papoutsis,
  B.~L. Saux, G.~Camps-Valls, and X.~X. Zhu, ``Neural plasticity-inspired
  multimodal foundation model for earth observation,'' \emph{arXiv preprint
  arXiv:2403.15356}, 2024.

\bibitem{tseng2026olmoearth}
G.~Tseng, Y.~Zhang, F.~Bastani, H.~Herzog, J.~Redmon, H.~Sablon, P.~Wolters,
  A.~Shah, P.~A. Johnson, C.~Wilhelm, and P.~Beukema, ``Olmoearth v1.1: A more
  efficient family of olmoearth models,'' \emph{arXiv preprint
  arXiv:2605.20804}, 2026.

\bibitem{jakubik2025terramind}
J.~Jakubik, F.~Yang, B.~Blumenstiel, E.~Scheurer, R.~Sedona, S.~Maurogiovanni,
  J.~Bosmans, N.~Dionelis, V.~Marsocci, N.~Kopp \emph{et~al.}, ``Terramind:
  Large-scale generative multimodality for earth observation,'' in
  \emph{Proceedings of the IEEE/CVF International Conference on Computer
  Vision}, 2025, pp. 7383--7394.

\bibitem{simeoni2025dinov3}
O.~Sim{\'e}oni, H.~V. Vo, M.~Seitzer, F.~Baldassarre, M.~Oquab, C.~Jose,
  V.~Khalidov, M.~Szafraniec, S.~Yi, M.~Ramamonjisoa \emph{et~al.}, ``Dinov3,''
  \emph{arXiv preprint arXiv:2508.10104}, 2025.

\bibitem{he2016resnet}
K.~He, X.~Zhang, S.~Ren, and J.~Sun, ``Deep residual learning for image
  recognition,'' in \emph{Proceedings of the IEEE Conference on Computer Vision
  and Pattern Recognition}, 2016, pp. 770--778.

\bibitem{liu2021swin}
Z.~Liu, Y.~Lin, Y.~Cao, H.~Hu, Y.~Wei, Z.~Zhang, S.~Lin, and B.~Guo, ``Swin
  transformer: Hierarchical vision transformer using shifted windows,'' in
  \emph{Proceedings of the IEEE/CVF International Conference on Computer
  Vision}, 2021, pp. 10\,012--10\,022.

\bibitem{dosovitskiy2020vit}
A.~Dosovitskiy, L.~Beyer, A.~Kolesnikov, D.~Weissenborn, X.~Zhai,
  T.~Unterthiner, M.~Dehghani, M.~Minderer, G.~Heigold, S.~Gelly \emph{et~al.},
  ``An image is worth 16x16 words: Transformers for image recognition at
  scale,'' in \emph{International Conference on Learning Representations},
  2021.

\bibitem{howard2019mobilenet}
A.~Howard, M.~Sandler, G.~Chu, L.-C. Chen, B.~Chen, M.~Tan, W.~Wang, Y.~Zhu,
  R.~Pang, V.~Vasudevan \emph{et~al.}, ``Searching for {MobileNetV3},'' in
  \emph{Proceedings of the IEEE/CVF International Conference on Computer
  Vision}, 2019, pp. 1314--1324.

\bibitem{hensman2015scalable}
J.~Hensman, A.~Matthews, and Z.~Ghahramani, ``Scalable variational gaussian
  process classification,'' in \emph{Artificial intelligence and
  statistics}.\hskip 1em plus 0.5em minus 0.4em\relax PMLR, 2015, pp. 351--360.

\bibitem{rs15174138}
M.~Czerkawski, R.~Atkinson, C.~Michie, and C.~Tachtatzis,
  ``Satellitecloudgenerator: Controllable cloud and shadow synthesis for
  multi-spectral optical satellite images,'' \emph{Remote Sensing}, vol.~15,
  no.~17, p. 4138, 2023.

\bibitem{foi2008practical}
A.~Foi, M.~Trimeche, V.~Katkovnik, and K.~Egiazarian, ``Practical
  poissonian-{Gaussian} noise modeling and fitting for single-image raw-data,''
  \emph{IEEE Transactions on Image Processing}, vol.~17, no.~10, pp.
  1737--1754, 2008.

\bibitem{sieberth2014motion}
T.~Sieberth, R.~Wackrow, and J.~H. Chandler, ``Motion blur disturbs -- the
  influence of motion-blurred images in photogrammetry,'' \emph{The
  Photogrammetric Record}, vol.~29, no. 148, pp. 434--453, 2014.

\bibitem{zhu2022blind}
B.~Zhu, Q.~Lv, Y.~Yang, X.~Sui, Y.~Zhang, Y.~Tang, and Z.~Tan, ``Blind
  deblurring of remote-sensing single images based on feature alignment,''
  \emph{Sensors}, vol.~22, no.~20, p. 7894, 2022.

\bibitem{el-yaniv2010foundations}
R.~El-Yaniv and Y.~Wiener, ``On the foundations of noise-free selective
  classification,'' \emph{Journal of Machine Learning Research}, vol.~11,
  no.~5, 2010.

\bibitem{geifman2017selective}
Y.~Geifman and R.~El-Yaniv, ``Selective classification for deep neural
  networks,'' in \emph{Advances in Neural Information Processing Systems
  (NeurIPS)}, vol.~30, 2017.

\bibitem{naeini2015ece}
M.~P. Naeini, G.~Cooper, and M.~Hauskrecht, ``Obtaining well calibrated
  probabilities using {Bayesian} binning,'' in \emph{Proceedings of the AAAI
  Conference on Artificial Intelligence}, 2015, pp. 2901--2907.

\bibitem{gneiting2007strictly}
T.~Gneiting and A.~E. Raftery, ``Strictly proper scoring rules, prediction, and
  estimation,'' \emph{Journal of the American Statistical Association}, vol.
  102, no. 477, pp. 359--378, 2007.

\bibitem{brier1950verification}
G.~W. Brier, ``Verification of forecasts expressed in terms of probability,''
  \emph{Monthly Weather Review}, vol.~78, no.~1, pp. 1--3, 1950.

\bibitem{kendall1945treatment}
M.~G. Kendall, ``The treatment of ties in ranking problems,''
  \emph{Biometrika}, vol.~33, no.~3, pp. 239--251, 1945.

\bibitem{kornblith2019similarity}
S.~Kornblith, M.~Norouzi, H.~Lee, and G.~Hinton, ``Similarity of neural network
  representations revisited,'' in \emph{International conference on machine
  learning}.\hskip 1em plus 0.5em minus 0.4em\relax PMlR, 2019, pp. 3519--3529.

\bibitem{nguyen2020wide}
T.~Nguyen, M.~Raghu, and S.~Kornblith, ``Do wide and deep networks learn the
  same things? uncovering how neural network representations vary with width
  and depth,'' \emph{arXiv preprint arXiv:2010.15327}, 2020.

\bibitem{ding2021grounding}
F.~Ding, J.-S. Denain, and J.~Steinhardt, ``Grounding representation similarity
  through statistical testing,'' \emph{Advances in Neural Information
  Processing Systems}, vol.~34, pp. 1556--1568, 2021.

\bibitem{mann1947test}
H.~B. Mann and D.~R. Whitney, ``On a test of whether one of two random
  variables is stochastically larger than the other,'' \emph{The Annals of
  Mathematical Statistics}, vol.~18, no.~1, pp. 50--60, 1947.

\bibitem{oquab2023dinov2}
M.~Oquab, T.~Darcet, T.~Moutakanni, H.~Vo, M.~Szafraniec, V.~Khalidov,
  P.~Fernandez, D.~Haziza, F.~Massa, A.~El-Nouby \emph{et~al.}, ``{DINOv2}:
  Learning robust visual features without supervision,'' \emph{Transactions on
  Machine Learning Research}, 2024.

\bibitem{yang2024generalized}
J.~Yang, K.~Zhou, Y.~Li, and Z.~Liu, ``Generalized out-of-distribution
  detection: A survey,'' \emph{International Journal of Computer Vision}, vol.
  132, no.~12, pp. 5635--5662, 2024.

\bibitem{lee2018simple}
K.~Lee, K.~Lee, H.~Lee, and J.~Shin, ``A simple unified framework for detecting
  out-of-distribution samples and adversarial attacks,'' in \emph{Advances in
  Neural Information Processing Systems}, vol.~31, 2018, pp. 7167--7177.

\bibitem{sun2022knnood}
Y.~Sun, Y.~Ming, X.~Zhu, and Y.~Li, ``Out-of-distribution detection with deep
  nearest neighbors,'' in \emph{Proceedings of the 39th International
  Conference on Machine Learning}, 2022, pp. 20\,827--20\,840.

\bibitem{xiao2018unified}
T.~Xiao, Y.~Liu, B.~Zhou, Y.~Jiang, and J.~Sun, ``Unified perceptual parsing
  for scene understanding,'' in \emph{Proceedings of the European Conference on
  Computer Vision (ECCV)}, 2018, pp. 418--434.

\bibitem{chen2018encoder}
L.-C. Chen, Y.~Zhu, G.~Papandreou, F.~Schroff, and H.~Adam, ``Encoder-decoder
  with atrous separable convolution for semantic image segmentation,'' in
  \emph{Proceedings of the European Conference on Computer Vision (ECCV)},
  2018, pp. 801--818.

\end{thebibliography}

\begin{IEEEbiographynophoto}{Nils Lehmann}
(n.lehmann@tum.de) is pursuing his Ph.D.\ degree at the Chair of Data Science in Earth Observation, Technical University of Munich (TUM), Germany. His research focuses on uncertainty quantification, evaluation of EO models, and practical applications for Earth observation data.
\end{IEEEbiographynophoto}

\begin{IEEEbiographynophoto}{Jakob Gawlikowski}
(jakob.gawlikowski@dlr.de) is with the DLR Remote Sensing Technology Institute, where he is leading the team Trustworthy AI4EO. His research focuses on the trustworthiness of machine learning approaches, with particular emphasis on data and model validation, uncertainty quantification (UQ), explainable AI (xAI), and multi-modal learning setups.
\end{IEEEbiographynophoto}

\begin{IEEEbiographynophoto}{Burak Ekim}
(burake@mit.edu) 
Burak Ekim is currently a postdoctoral associate at the Computer Science and Artificial Intelligence Laboratory (CSAIL) of the Massachusetts Institute of Technology (MIT). Burak works on machine learning methods for environmental conservation, computational ecology, and disaster response using geospatial and citizen science data, with a focus on robustness, active learning, and scientific discovery.

\end{IEEEbiographynophoto}

\begin{IEEEbiographynophoto}{Isaac Corley}
(isaac.corley@taylorgeospatial.org) is Director of Research at Taylor Geospatial where he leads efforts to bring geospatial AI from research to global-scale products for all to benefit from. He received his Ph.D.\ in Electrical Engineering from the University of Texas at San Antonio (UTSA) in 2023. His current research interests consist of earth embeddings, computer vision applications for Earth observation, and improving evaluation of geospatial foundation models to align with realistic stakeholder use-cases.
\end{IEEEbiographynophoto}

\begin{IEEEbiographynophoto}{Xiao Xiang Zhu}
(xiaoxiang.zhu@tum.de) is the Chair Professor for Data Science in Earth Observation at the Technical University of Munich (TUM), Munich, Germany, where she also serves as Director of the Munich Data Science Institute (MDSI). She is currently a Visiting AI Professor at the European Space Agency's $\phi$-lab, Frascati, Italy. Since 2022, she leads the national center of excellence ML4Earth (ml4earth.de), funded by the German Space Agency. Her research interests include Earth observation and remote sensing, signal processing, machine learning, and data science, with applications to global urbanization, the UN Sustainable Development Goals, and climate change. More information
is available at https://www.asg.ed.tum.de/sipeo
\end{IEEEbiographynophoto}

\end{document}